\documentclass[journal,twoside]{IEEEtran}

\usepackage{ifpdf}
\ifCLASSINFOpdf
  \usepackage[pdftex]{graphicx}
 \DeclareGraphicsExtensions{.pdf,.jpeg,.png,.jpg}
\else
  \usepackage[dvips]{graphicx}
  \DeclareGraphicsExtensions{.eps}
\fi
\usepackage{epstopdf}
\usepackage{calc}
\usepackage{mathptmx} 
\usepackage[cmex10]{amsmath}
\usepackage{amssymb}

\usepackage{listings}
\usepackage[lined,linesnumbered,noend,figure]{algorithm2e}
\usepackage{dblfloatfix}
\usepackage{url}
\usepackage[absolute]{textpos}

\begin{document}
\clearpage \newpage
\title{Data Generators for Learning Systems\\  Based on RBF Networks }
\author{Marko Robnik-\v{S}ikonja 
  \thanks{This work was supported by the Slovenian Research Agency (ARRS) through research programme P2-0209 and FP7 European Commission  project HBP (Human Brain Project, grant agreement no. 604102).}
  \thanks{M. Robnik-\v{S}ikonja is with University of Ljubljana, 
  Faculty of Computer and Information Science, Slovenia
  (Marko.Robnik@fri.uni-lj.si)}
   }

\begin{textblock}{14}(1.3,0.5)
\tt{\noindent  Published in:\\
\centering	IEEE Transaction on Neural Networks and Learning Systems, 27(5):926-938, 2016}
\end{textblock}

\maketitle
\begin{abstract}
There are plenty of problems where the data available is scarce and expensive. We propose a generator of semi-artificial data with similar properties to the original data which enables development and testing of  different data mining algorithms and optimization of their
parameters. The gene\-rated data allow a large scale experimentation and simulations without danger of overfitting. The proposed generator is based on RBF networks, which learn sets of Gaussian kernels. These Gaussian kernels  can be used in a generative mode to generate new data from the same distributions. To assess quality of the generated data we evaluated the statistical properties of the generated data, structural similarity and predictive similarity using supervised and unsupervised learning techniques. To determine usability of the proposed generator we conducted a large scale evaluation using 51 UCI data sets. The results show a considerable similarity between the original and generated data and indicate that the method can be useful in several development and simulation  scenarios. 
We analyze possible improvements in classification performance by adding different amounts of generated data to the training set, performance on high dimensional data sets, and conditions when the proposed approach is successful.
\end{abstract}
\section{Introduction}
\IEEEPARstart{O}{ne} of technological challenges faced by data analytics is an enormous amount of data. This challenge is well known and a term "big data" was coined with the purpose to bring attention to it and to develop new solutions. However, in many important application areas the excess of data is not a problem, quite the opposite, there is actually not enough data available. There are several reasons for this, the data may be inherently scarce (rare diseases, faults in complex systems, rare grammatical structures...), difficult to obtain (due to proprietary systems, confidentiality of business contracts, privacy of records), expensive (obtainable with expensive equipment, requiring significant investment of human or material resources), or the distribution of the events  of interests is highly imbalanced (fraud detection, outlier detection, distributions with long tails). For machine learning approaches the lack of data causes problems in model selection, reliable performance estimation, development of
 specialized algorithms, and tuning of learning model parameters. While certain problems caused by scarce data are inherent to underrepresentation of the problem and cannot be solved, some aspects can be alleviated by generating artificial data similar to the original one. For example, similar artificial data sets can be of great help in tuning the parameters, development of specialized solutions, simulations, and imbalanced problems, as they prevent overfitting of the original data set, yet allow sound comparison of different approaches.

Generating new data similar to a general data set is not an easy task. If there is no background knowledge available about the problem, we have to use the precious data we posses to extract some of its properties and generate new semi-artificial data with similar properties. Weather this is acceptable in the context of the problem is not a matter of the proposed approach, and we assume that one can afford to set aside at least small part of the data for this purpose. This data may not be lost for modeling, but we shall be aware of extracted properties when considering possibility of overfitting. 

The approaches used in existing data generators are limited to low dimensional data (up to 6 variables) or assume certain probability distribution, mostly normal; we review existing approaches in Section 2.  Our method is limited to classification problems but we do not assume any fixed data distribution. We first construct a RBF network prediction model.  RBF networks consist of several Gaussian kernels, which estimate local probability density from training instances.
Due to properties of Gaussian kernels  (discussed in Section 3), the learned kernels can be used in a generative mode to produce new data. In this way we overcome limitation to low dimensional spaces and capture complexity of real-world data sets. We show that our approach can be successfully used for data sets with several hundred  attributes and also with mixed data (numerical and categorical). 

The paper is organized as follows. In Section 2 we review existing work on generating semi-artificial data. In Section 3 we present RBF neural networks and properties which allow us to generate data based on them. In Section 4 we present the actual implementation and explain details on handling nominal and numeric data.  
In Section 5 we present the quality of the generated data and the working conditions of the proposed method. 
We analyze behavior of classifiers when new data is added to the training set and shortly present the use of the generator for benchmarking in cloud based big data analytics tool. In Section 6 we conclude with a summary, critical analysis and ideas for further work.

\section{Related work}
We review only general approaches to data generation and do not cover methods specific for a certain problem or a class of problems.
The largest group of existing data generators is based on an assumption about specific probability distribution the generated data shall be drawn from. Most computational libraries and tools contain
random number generators for univariate data drawn from standard distributions.
For example, R system \cite{R} supports uniform, normal, log-normal, Student's t, F,  Chi-squared, Poisson, exponential, beta, binomial, Cauchy, gamma, geometric, hypergeometric, multinomial, negative binomial, and Weibull distribution. Additional univariate distribution-based random number generators are accessible through add-on packages. If we need univariate data from these distributions based on empirical data, we fit the parameters of the distributions and then use the obtained parameters to generate new data. For example, R package MASS \cite{Venables02} provides function \textit{fitdistr} to obtain the parameters of several univariate distributions. 

New samples can be constructed from probability density estimates obtained with kernel density estimation functions. For example, \cite{Li06} uses Gaussian kernels to estimate univariate densities of attributes in manufacturing process data set and generates new virtual instances which are added to the original small data set to improve learning. This approach works only when adding a small number of instances, fails for larger samples, and is tested on a single problem. Virtual samples are generated also in \cite{Yang11}, where authors perturb original instances 
by adding normally distributed noise. These approaches work for low dimensional problems. 

Random vector generators are mostly based on multivariate t and normal distributions, and are effective with up to 6 variables. Simulating data from multivariate normal distribution is possible via decomposition
of a given symmetric positive definite matrix $\Sigma$ containing variable covariances. Using the decomposed matrix and a sequence of univariate normally distributed random variables one can generate data from multivariate normal distribution as discussed in Section \ref{sec:rbfGenerator}. The approach proposed in this paper relies on the multivariate normal distribution data generator but does not assume that the whole data set is normally distributed. Instead it finds subspaces which can be successfully approximated with Gaussian kernels and use extracted distribution parameters to generate new data in proportion with the requirements. 

To generate data from nonnormal multivariate distribution several transformational approaches have been proposed which 
start by generating data from a multivariate normal distribution and than transform the data to the desired final distribution. 
For example, \cite{Ruscio08} proposes an iterative approximation scheme. 
In each iteration the approach generates a multivariate normal data that is subsequently replaced with the nonnormal data sampled
from the specified target population. After each iteration discrepancies between the generated and desired correlation
matrices are used to update the intermediate correlation matrix. A similar approach for ordinal data is proposed in \cite{Ferrari12}.
Transformational approaches are limited to low dimensional spaces, where the covariance matrix capturing data dependencies can be successfully estimated. In contrast, our method is not limited to low dimensional spaces. The problem space is split into subspaces, 
where dependencies are more clearly expressed and captured. 

Kernel density estimation is a method, which estimates the probability density function of a random variable with a  kernel function. The inferences about the population are made based on a finite data sample. Several approaches for kernel based parameter estimation exist. The most frequently used kernels are Gaussian kernels. These methods  are intended for low dimensional spaces with up to 6 variables \cite{Hardle00}.

An interesting approach to data simulations uses copulas \cite{Nelsen99}.  A copula is a multivariate probability distribution, for which the marginal probability distribution of each variable is uniform. Copulas can be estimated from empirical observations and describe dependencies between random variables.  They are based on Sklar's theorem stating that any multivariate joint distribution can be written with univariate marginal distribution functions and a copula, which describes the dependence structure between variables.
To generate new data one has to first select the correct copula family, estimate the parameters of the copula, and then generate the data.
The process is not trivial and requires in-depth knowledge of the data being modeled. In principle the number of variables used in a copula is not limited, but in practice a careful selection of appropriate attributes and copula family is required \cite{Bandara11,Mair12}. Copulas for both numeric and categorical data exist, but not for mixed types, whereas our approach is not limited in this sense and is completely automatic.

An evolutionary approach to generating new data is presented in \cite{Meraviglia06}. The approach uses subsamples of a given data set to construct several neural networks of different characteristics (this shall assure that different features of the data set are captured). Each of the generated networks is used together with a separate run of a genetic algorithm to assess if the population of instances in the genetic algorithm has converged to a data set similar to a separate testing set. The similarity between generated data and the testing set is estimated using classification performance of the neural networks.
The approach was tested on a social survey data set and showed good classification-based similarity, but notable differences in pairwise distribution-based similarity of individual features were detected. The authors observe that dependencies between dependent and independent variables are reflected in the generated data, but dependencies between independent variables are mostly lost. As our approach uses the same model for both learning model and generation of new instances, attribute dependencies are preserved. We use extended notion of data similarity, which besides the classification performance includes also clustering and attribute statistics. Our approach is evaluated on several data sets and shows robust performance.

\markboth{M. Robnik-\v{S}ikonja: Data generators for learning systems based on {RBF} networks}{M. Robnik-\v{S}ikonja: Data generators for learning systems based on {RBF} networks }

\section{RBF networks}
\begin{figure}[t]
\centerline{\includegraphics[width= 0.47 \hsize]{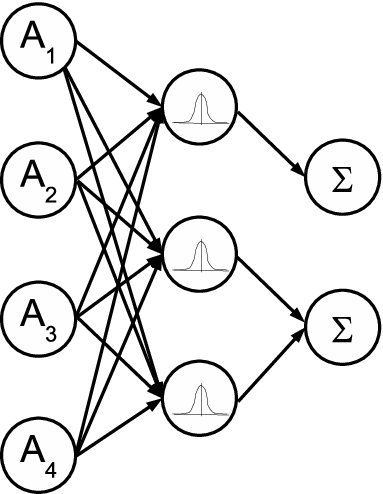}} 
\caption{Structure of RBF-DDA network for classification problem with 4 attributes, 3 hidden units, and a binary class.} 
 \label{fig:RBF-DDA}
\end{figure}

The RBF (Radial Basis Functions) networks were proposed in 1989 as a function approximation tool \cite{Moody89}.  They use locally tuned processing units, mostly Gaussian kernels. Today RBF networks are one of the standard artificial neural network learning methods and are included in several learning packages, for example SNNS used in this work \cite{Zell95}. Due to their success and versatility their development is still ongoing (see for example  \cite{Yao12,Zhang13}). 
A typical RBF network consists of three layers (see Fig. \ref{fig:RBF-DDA} for an illustration). The input layer has $a$ input units, corresponding to input features. The hidden layer contains $k$ kernel functions. The output layer consists of a single unit in case of regression or as many units as there are output classes in case of classifications. We assume a classification problem described with $n$ pairs of $a-$dimensional training instances 
$({\bf x}_i,  y_i)$, where ${\bf x}_i \in \Re^a$ and  $y_i$ is one of class labels $1,2... C$. Hidden units in RBF network estimate the probability of each class $y_u$: 
$$
p(y_u|{\bf x}) = \sum_{j=1}^k c_j h_j ({\bf x}).
$$
The weights  $c_j$ are multiplied by radial basis functions $h_j$, which are usually Gaussian kernels: 
$$
h_j ({\bf x})  = \exp(-\frac{||{\bf x} - {\bf t}_j||}{\sigma_j^2}).
$$
The vector  ${\bf t}_j$ represents the $j$-th center and $\sigma_j$ represents the width of a kernel.  The centers and kernel widths  have to be learned or set in advance. 
The kernel function $h$ is applied to the Euclidean distance between each center ${\bf t}_j$  and given instance  ${\bf x}$.  The kernel function has its maximum at zero distance from the center, while the activation is close to zero for instances which are further away from the center.

Most algorithms used to train RBF networks require a fixed architecture in which the number of units in the hidden layer must be determined before the training starts.
To avoid manual setting of this parameter and to automatically learn kernel centers ${\bf t}_j$, weights $c_j$, and standard deviations $\sigma_j$, several solutions have been proposed \cite{Reilly82,Berthold95,Zhang13,Yao12}, among them RBF  with Dynamic Decay Adjustment (DDA)\cite{Berthold95} with a stable and well-tested implementation which we use in this work. The RBF DDA builds a network 
by incrementally adding  RBF units. Each unit encodes instances of only one class. During the process of adding new units the kernel widths $\sigma_j$ are dynamically adjusted (decayed) based on information about the neighbors. 
RBFs trained with the DDA algorithm often achieve classification accuracy comparable to multi layer perceptrons  \cite{Berthold95, Zell95}.

An example of RBF-DDA network for classification problem with 4 features and a binary class is presented in Fig. \ref{fig:RBF-DDA}. The hidden layer of RBF-DDA network contains Gaussian units, which are added during
training. The input layer is fully connected to the hidden layer. The output layer consists of one unit for each possible class. 
Each hidden unit encodes instances of one class and is therefore connected to exactly one output unit.
 For classification of a new instance a winner-takes-all approach is used, i.e. the output unit with the highest activation determines the class value. 

Our data generator uses the function \textit{rbfDDA} implemented in R package RSNNS \cite{Bergmeir12} which is a R port of SNNS
software \cite{Zell95}. The implementation
uses two parameters: a positive threshold $\Theta^+$ and a negative threshold $\Theta^-$. 
The two thresholds define the upper and lower bound for the activation of training instances. Default values of thresholds are $\Theta^+=0.4$ and $\Theta^-=0.2$.  The thresholds define a safety area where no other center of a conflicting class is allowed. In this way a good separability of classes is achieved. In addition, each training instance has to be in the inner circle of at least one center of the correct class.

\section{Data generator }
\label{sec:rbfGenerator}
The idea of the proposed data generation scheme is to extract Gaussian kernels from the learned RBF-DDA network and generate data from each of them in proportion to the desired class value distribution. When  class distribution different  from the empirically observed one is desired, the distribution can be specified as an input parameter.
 
A notable property of Gaussian kernels is their ability not to be used only as discriminative models but also as generative models. 
To generate data from multivariate normal distribution $N(\mu, \Sigma)$ one can exploit the following property of multivariate Gaussian distributions:  
\begin{equation}
{\rm if}~~ {\bf X} \sim N(\mu, \Sigma), ~{\rm then}~
{\bf Y} = A{\bf X} + {\bf b} \sim N(A\mu + {\bf b},~ A\Sigma A^T).
\label{eq:linN}
\end{equation}

When we want to simulate multidimensional ${\bf Y} \sim N(\mu, \Sigma)$,  for a given symmetric positive definite matrix $\Sigma$, we first construct a sample ${\bf X} \sim N({\bf 0},1)$ of the same dimensionality. The ${\bf X} \sim N({\bf 0},1)$ can easily be constructed using independent variables  $X_i \sim N(0,1)$. Next we decompose $\Sigma = A A^T$ (using Cholesky or eigenvalue decomposition). With the obtained matrix  $A$  and {\bf X} we use Equation (\ref{eq:linN}) to get
\begin{equation}
{\bf Y } = A {\bf X} + \mu  \sim N(A {\bf 0} + \mu, A 1A^T) = N(\mu, \Sigma).
\label{eq:newN}
\end{equation}
In our implementation we use function \textit{mvrnorm} from R package MASS \cite{Venables02}, which decomposes covariance matrix $\Sigma$ with eigenvalue decomposition due to better stability \cite{Ripley87}. 

\subsection{Construction of generator}
\label{sec:rbfGen}
The pseudo code of the proposed generator is given in Fig. \ref{alg:rbfGen}. The inputs to the generator are the available data set and two parameters. The parameter $minW$ controls the minimal acceptable kernel weight. The weight of the kernel is defined as the number of training instances which achieve maximal activation with that kernel (and therefore belong to that kernel). All the learned kernels with weight less than $minW$ are discarded by data generator to prevent overfitting of the training data. The boolean parameter $nominalAsBinary$ controls the treatment of nominal attributes as described in Section \ref{sec:rbfPrepare}.

Due to specific requirements of RBF-DDA algorithm, the data has to be preprocessed  (line 2 in Fig. \ref{alg:rbfGen}). The preprocessing includes normalization of attributes to $[0,1]$  and preparation of nominal attributes (see Section \ref{sec:rbfPrepare}). Function rbfPrepareData returns normalized data $D$ and normalization parameters  $N$ which are used later when generating new instances.
The learning algorithm takes the preprocessed data and returns the classification model $M$ consisting f Gaussian kernels (line 3). 
We store the obtained parameters of the Gaussian kernels, namely their centers ${\bf t}$, 
weights $w$, and class values $c$  (lines 4, 5, and 6).  The kernel weight $w_k$ equals the proportion of training instances activated by the $k$-th Gaussian unit.
The class value $c_k$ of the unit corresponds to the output unit connected to the Gaussian unit $k$ (see Fig. \ref{fig:RBF-DDA} for illustration). The extracted information is sufficient to generate new data, however, there are several
practical considerations which have to be taken into account if one is to generate new data similar to the original one.

\begin{algorithm*}[t!!]
\DontPrintSemicolon
\SetCommentSty{textsf}
\SetKwFunction{rbfGen}{rbfGen}
\SetFuncSty{textrm}
\KwIn{data set $D=\{({\bf x}_i,  y_i)_{i=1}^n\}$, parameters $minW$, $nominalAsBinary$}
\KwOut{a list $L$ of Gaussian kernels and  a list $N$ of attribute normalization parameters}
\BlankLine
{\bf Function} \rbfGen{$D$, $minW$,  $nominalAsBinary$}\;
\tcp*[l]{preprocess the data to get $[0,1]$ normalized data $D$ and normalization parameters $N$}
$(D, N) \gets $ rbfPrepareData($D$, $nominalAsBinary$)\;
$M \gets$  rbfDDA($D$) \tcp*[l]{learn RBF model consisting of $k$ kernels }
\ForEach{{\rm kernel} $k \in M$} {
   \If(\tcp*[h]{store only kernels with sufficient weight}){$w_k \ge minW$}{
     $L_k \gets  ({\bf t}_k, w_k, c_k)$\tcp*[l]{store center,  weight, and class}
  }
}
\For(\tcp*[h]{find activation unit of each instance}){$i \in 1\dots n $}
 { $z_i \gets \arg\max_{k\in M}  \exp(\frac{-||{\bf x_i} - {\bf t}_k||}{\sigma_k^2})$ \;}
\ForEach(\tcp*[h]{estimate empirical kernel width}){{\rm kernel} $k \in M$} {
   $\Sigma_k \gets $  std($\{{\bf x}_i; z_i=k\}$) \tcp*[l]{compute spread on matching instances for each dimension}
   $L_k \gets L_k \cup \Sigma_k$ \tcp*[l]{ add $\Sigma_k$ to output list item $L_k$}
}
\KwRet{(L, N)}
\caption{The pseudo code of creating a RBF based data generator.}
\label{alg:rbfGen}
\end{algorithm*}

The task of RBF-DDA is to discriminate between instances with different class values, therefore 
during the learning phase widths of kernels are set  in such a way that majority of instances are activated by exactly one kernel. Narrow widths of  learned kernels therefore prevent overlapping of competing classes. For the purpose of generating new data 
widths of kernels shall be wider, or we would only generate instances in the near proximity of kernel centers i.e. existing training instances. The approach we adopted is to take the training instances that activate the particular kernel (lines 7 and 8) 
and estimate their empirical variance  in each dimension (lines 9, 10, and 11) which is later, in the generation phase, used as the width of the Gaussian kernel. The $\Sigma$ matrix extracted from the network is diagonal, with elements presenting the spread of training instances in each dimension.
The algorithm returns the data generator consisting of the list of kernel parameters $L$ and data normalization parameters $N$ (line 12).

\subsection{Preprocessing the data}
\label{sec:rbfPrepare}
Function rbfPrepareData performs three tasks: it imputes missing values,   prepares nominal attributes, and normalizes data. The pseudo code of data preprocessing is in Fig. \ref{alg:rbfPrepare}. 

\begin{algorithm*}[ht]
\DontPrintSemicolon
\SetCommentSty{textsf}
\SetKwFunction{rbfPrepare}{rbfPrepareData}
\SetFuncSty{textrm}
\SetKwIF{If}{ElseIf}{Else}{if}{}{elif}{else}{}%
\KwIn{data set $D=\{({\bf x}_i,  y_i)_{i=1}^n\}$, parameter $nominalAsBinary$}
\KwOut{preprocessed data $D'$, a list $T$ with information about attribute transformations }
\BlankLine
{\bf Function} \rbfPrepare{D, nominalAsBinary}\;
\For(\tcp*[h]{preprocessing of attributes}){$j \in 1\dots a $}   {
   $D'({\bf x}_{.j}) \gets$ imputeMissing(${\bf x}_{.j}$)\tcp*[l]{imputation of missing values}
  \If(\tcp*[h]{encode nominal attributes}) {({\rm isNominal}(${\bf x}_{.j}$))}{
     \If{($nominalAsBinary$)}{
        $D' \gets$  encodeBinary(${\bf x}_{.j}$)
     }
     \Else{
        $D' \gets$  encodeInteger(${\bf x}_{.j}$)
     }
}
 $D'({\bf x}_{.j}) \gets \frac{{\bf x}_{.j} - \min {\bf x}_{.j}}{\max {\bf x}_{.j} - \min {\bf x}_{.j}}$ 
         \tcp*[l]{ normalize attributes to $[0,1]$}
  \tcp*[l]{store normalization and encoding parameters}
  $T_j \gets$ ($\min {\bf x}_{.j},\ \max {\bf x}_{.j} - \min {\bf x}_{.j}$,  encoding(${\bf x}_{.j}$) )  
}
 $D' \gets$ replace class $y$ with its binary encodings\;
\KwRet{(D', T)}
\caption{Preprocessing data for RBF-DDA algorithm; $x_{.j}$ stands for values of attribute $j$.}
\label{alg:rbfPrepare}
\end{algorithm*}
%
The function \textit{rbfDDA} in R does not accept missing values, so we have to impute them (line 3). While several advanced imputation strategies exist, we, due to efficiency reasons, use simple median based imputation for numeric attributes, while for nominal attributes we impute the most frequent category. 
For a specific data set and intended classifier a better choice may exist \cite{Farhangfar08} and  R package \lstinline!missForest! \cite{Stekhoven12} is a good default choice for mixed type data.

Gaussian kernels are defined only for numeric attributes, therefore \textit{rbfDDA} treats all attributes, including nominal, as numeric. Each nominal attribute is converted to numeric type (lines 4-8). We can simply assigning each category a unique integer from 1 to the number of categories (line 8). This may be problematic as this transformation has established an order of categorical values in the converted attribute, inexistent in the original attribute. For example, for attribute $Color$ the categories $\{ red, green, blue \}$ are converted into values $\{1, 2, 3\}$, respectively, meaning that the category $red$ is now closer to $green$ than to $blue$. To solve this problem we use the binary parameter $nominalAsBinary$ (line 5) and encode nominal attributes with several binary attributes when this parameter is set to \textit{true} (line 6). Nominal attributes with more than two categories are encoded with the number of binary attributes equal to the number of categories. Each category is encoded by one binary attribute. If the value of the nominal attribute equals the given category, the value of the corresponding binary attribute is set to 1, while the values of the other encoding binary attributes are set to 0. E.g., $Color$ attribute with three categories would be encoded with three binary attributes $C_{red}, C_{green}, C_{blue}$. If the value of the original attribute is $Color=green$ then the binary encoding of this value is  $C_{red}=0$, $C_{green}=1$, and $C_{blue}=0$. This binary encoding is required for class values (line 11).

The function \textit{rbfDDA} in R expects data to be normalized to $[0,1]$ interval (line 9). As we want to generate new data in the original, unnormalized form, we store the computed normalization parameters (line 10) and, together with attribute  encoding information, return them to the function \textit{rbfGen}.  

\subsection{Generating new data}
\label{sec:newdata}
Once we have the generator (produced by function \textit{rbfGen}), we can use it to generate new instances. By default the method generates 
new instances labeled with class values proportionally to the number of class values in the training set of the generator, but the user can specify the desired class distribution as a parameter vector $p$.

\begin{algorithm*}[t]
\DontPrintSemicolon
\SetCommentSty{textsf}
\SetKwFunction{rbfNewdata}{newdata}
\SetFuncSty{textrm}
\KwIn{$L$ - a list of Gaussian kernels, $T$ - an information about attribute normalization and encoding, $size$ - the number of instances to be generated, $p$ - a vector of the desired class distribution, $var$ - a parameter controlling the width of kernels, \textit{defaultSpread} - the width of the kernel if estimated width is 0 }
\KwOut{new data set $F=\{({\bf x}_i,  y_i)_{i=1}^{size}\}$}
\BlankLine
{\bf Function} \rbfNewdata{L, T, size, p, var, defaultSpread}\;
$D \gets \{\}$\tcp*[l]{create an empty temporary data set}
\ForEach{{\rm kernel} $k \in L$} {
   $g \gets \frac{w_k}{\sum_{i=1}^{|L|} w_i \cdot  I(c_i = c_k)} \cdot p_{c_k} \cdot size$\tcp*[l]{number of instances to generate with this kernel}
    \tcp*[l]{set kernel width }
   \lIf(\textsf{ }){(var={\rm "estimated")}}{ $\Sigma=\Sigma_k$ with zeros substituted by \textit{defaultSpread}}
   \lElseIf( \textsf{// heuristic rule}){(var={\rm "Silverman")}}{$\Sigma$ = silverman($\Sigma_k, n, a$) }
   $H \gets$ mvrnorm(n=g, mu=${\bf t}_k$, Sigma=$\Sigma$)\tcp*[l]{generate new data with kernel $k$}
  $H \gets$ makeConsistent($H, T_k$)\tcp*[l]{check and fix inconsistencies} 
  $H(y) \gets c_k$\tcp*[l]{assign class value from the kernel }
   $D \gets D \cup H$ \tcp*[l]{append generated data to $D$}
}
\For(\tcp*[h]{transform attributes back to original scales and encodings}){$j \in 1\dots a $}   {
  \If(\tcp*[h]{decode nominal attributes}) {($T_j$.{\rm nominal})}{
     \If{($T_j$.{\rm binaryEncoded})}{
        $F({\bf x}_{.j}) \gets$  decodeBinary($D({\bf x}_{.j})$, $T_j$)\;
     }
     \Else{
        $F({\bf x}_{.j}) \gets$  decodeInteger($D({\bf x}_{.j})$, $T_j$)\;
     }
}
 \Else{
$F({\bf x}_{.j}) \gets D({\bf x}_{.j})  \cdot T_j.{\rm span}  + T_j.{\rm min}$ \tcp*[l]{ denormalize numeric attributes}
}
}
\KwRet{F}
\caption{The pseudo code for creating new instances with RBF generator.}
\label{alg:rbfNewdata}
\end{algorithm*}

A data generator consists of a list $L$ with parameters describing Gaussian kernels and a list $T$ with information about attribute transformations. Recall that information for each kernel $k$ contains the location of kernel's center ${\bf t}_k$, weight of kernel $w_k$, class value $c_k$, and estimated standard deviation $\Sigma_k$. An input to function \textit{newdata} are also parameters $size$ specifying the number of instances to be generated, $p$ the desired distribution of class values, $var$ specifying a method  for setting the width of the kernels, and \textit{defaultSpread} as the substitute kernel width in case the estimated width in certain dimension is 0.

Function starts by creating an empty data set $D$ (line 2) and than generates instances with each of the kernels stored in the kernel list $L$ (lines 2-11).The weight $w_k$ of the kernel $k$, the desired class probability $p_k$, and the overall number of instances to be generated $size$ determine the number of instances $g$ to be generated with each kernel (line 4). The weight of each kernel is normalized with the sum of weights of the same class kernels  $\frac{w_k}{\sum_{i=1}^{|L|} w_i \cdot  I(c_i = c_k)}$, where $I()$ is an indicator function, which returns 1 if its argument is \textit{true} and 0 otherwise. The width of the kernel determines the spread of the generated values around the center. By default we use the spread estimated from the training data (line 5). Zero spread in individual dimensions is optionally replaced by the value of the parameter \textit{defaultSpread}.
To set kernel width it is also possible to use the generalization of Silverman's  rule of thumb for multivariate case (line 6), but in our experiments this did not work well for data sets with more than 4 attributes.

The data is generated by function \textit{mvrnorm} (line 7). The function takes as its input $g$, the number of instances to generate, the center of the kernel ${\bf t}_k$, and the covariance matrix $\Sigma_k$, which is diagonal in our case. Function exploits the property of Gaussian kernels from Equation (\ref{eq:linN}) and decomposes covariance matrix $\Sigma$ with eigenvalue decomposition. The generated data has to be checked for consistency (line 8), i.e., generated attribute values have to be in $[0,1]$ interval, nominal attribute values have to be rounded to the values encoding existing categories, etc. As some instances are rejected during this process in practice we generate 
more than $g$ instances with \textit{mvrnorm} but retain only the desired number of them. We assign class values to the generated  instances (line 9) and append them to $D$ (line 10).
When the data are generated we transform the generated instances back to the original scales and encodings. For each nominal attribute  we check its encoding (either as a set of binary attributes or as an integer), and transform it back to the original form (lines 11-18). Numeric attributes are denormalized and transformed back to the original scale using minimums and spans stored in $T$ (line 18).  The function returns the generated data set (line 19).

\subsection{Visual inspection of generated data}
\begin{figure*}[!t]
\centerline{
\includegraphics[width= 0.47 \hsize]{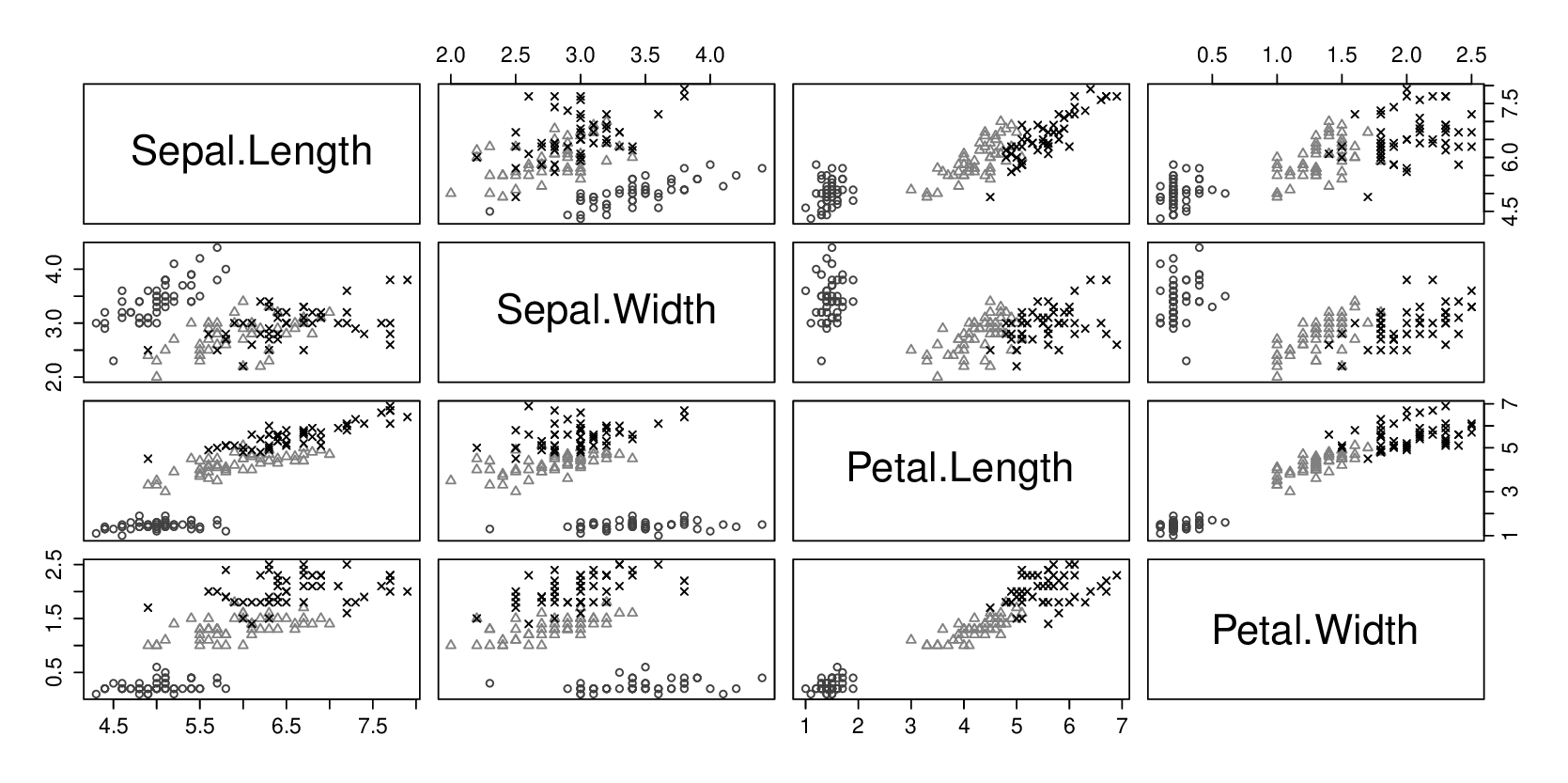} \hfil 
\includegraphics[width= 0.47 \hsize]{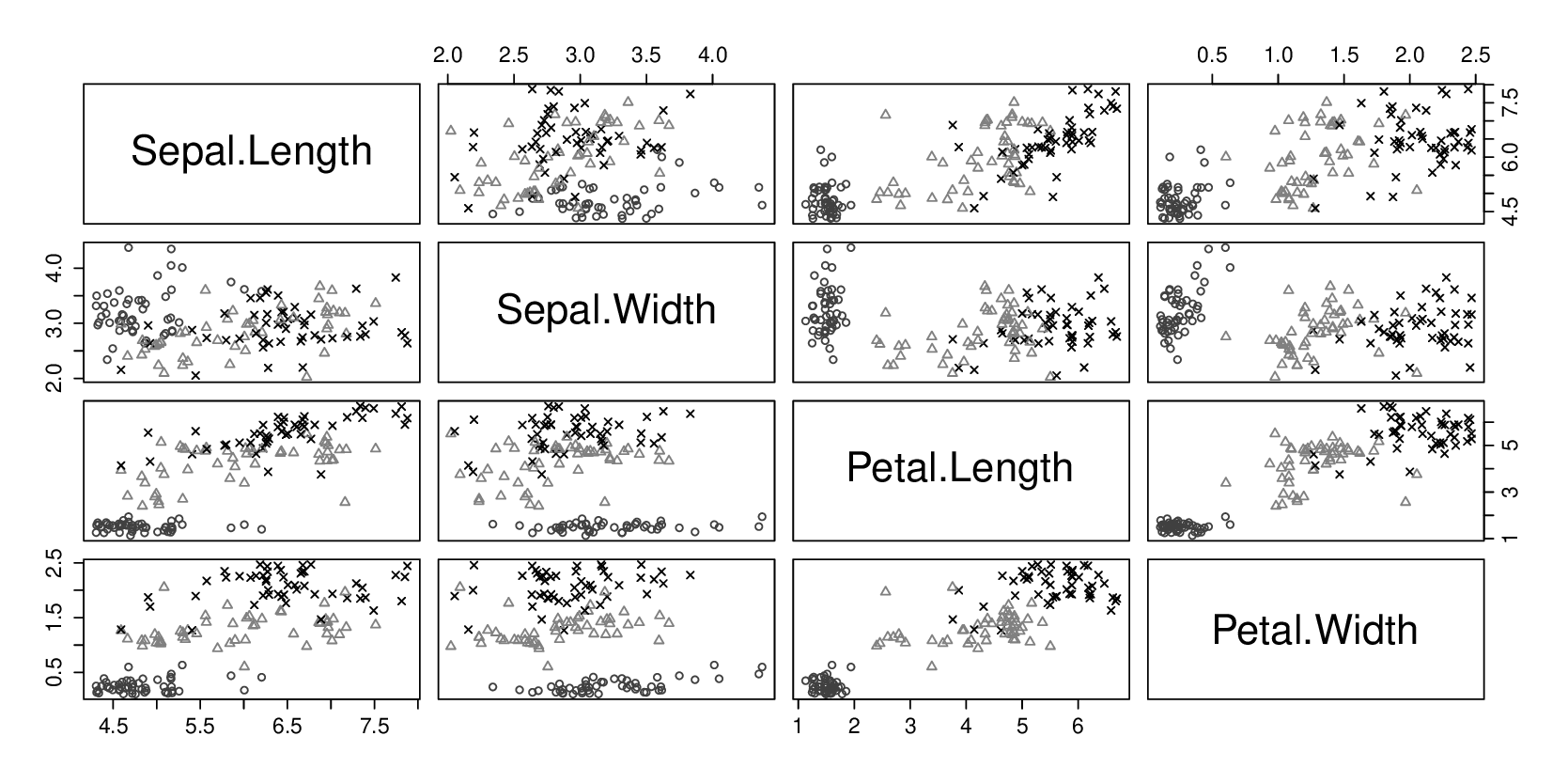} 
}
\caption{Scatter plots of original (left-hand side) and generated data (right-hand side) for Iris data set.} 
 \label{fig:iris}
\end{figure*}

As a demonstration of the generator we graphically present the generated data from a well known Iris data set,
which consists of 50 samples from each of three species of iris (\textit{Iris setosa, Iris virginica,} and \textit{Iris versicolor}). Four features were measured from each sample: the length and the width of the sepals and petals, in centimeters. The scatter plots of the original data set are shown on the left-hand side of Fig. \ref{fig:iris},
where class values are marked with different colors and different characters. The matrix shows dependencies between pairs of features, which are depicted on x- and y-axis, respectively.
The generator based on this data consists of 31 Gaussian units. We used it to generate 150  instances shown on the right-hand side of Fig. \ref{fig:iris}. The graphs show considerable similarity between matching pairs of scatter plots. Please note that the plots are just a visual demonstration of the generator and does not necessary prove its quality. The performance evaluation is a topic of the following Section.
\section{Evaluation}
To the best of our knowledge there exists no other universal data generator, that is, a generator capable of generating numerical and categorical data with large number of dimensions. We also do not compare our approach to generators constructed specifically for certain purpose (where our method is probably worse), but instead try to show its merit and limitations.
We verify if the proposed generator produces data  consistent with the original and 
try to determine working conditions of the generator: on which data set it works and where it fails, on what sort of problems it veritably reproduces the original and where it is less successful. 
We add generated data to training set to analyze its effect on predictive performance and describe the use of the generator in big data tool benchmarking. 

We performed a large scale empirical evaluation using 51 data sets from UCI repository \cite{UCI14} with great variability in the number of attributes, types of attributes and number of class values. 
Assuming that one would generate semi-artificial data when the number of original instances is small and also to keep the computational load of the evaluation low, we limited the number of original instances to be between 50 and 1000 (lower limit is necessary to assure sufficient data for both the generator and the testing set). Taking these conditions into account we  extracted 51 classification data sets from a collection of 92 data sets provided by R package readMLData \cite{readMLData}, which provides  a uniform interface for manipulation of UCI data sets. 

 For each data set a generator based on \textit{rbfDDA} classifier was constructed with function \textit{rbfGen} (Fig. \ref{alg:rbfGen}).
For all data sets we used parameters $minW=1$ and binary encoding of nominal attributes. The produced generator was used to generate the same number of instances as in the original data set and with the same distribution of classes using the function \textit{newdata} (Fig. \ref{alg:rbfNewdata}). The widths of  kernels were estimated from the training set by setting the parameter \textit{var=}"estimated" and \textit{defaultSpread=}0.05.

We compare the original and generated data sets using three types of comparisons.

\textit{1. Statistics of attributes:} we compare standard statistics of numeric attributes: 
mean, standard deviation, skewness, and kurtosis. 
Each numeric attribute is normalized to $[0, 1]$ to make comparison between attributes sensible for all statistics.
Input to each comparison are attributes from two data sets (original and generated). Comparison of attributes' statistics computed on both data sets is tedious, especially for data sets with large number of attributes. We therefore first compute statistics on attributes and then subtract statistics of the second data set from statistics of the first data set. To summarize the results we report only the median of absolute values of the difference for each of the statistics. 

To compare distributions of attribute values we use He\-llin\-ger distance for discrete attributes and KS test for numerical attributes.
 For numerical attributes we compute p-values of KS tests under the null hypothesis that attri\-bu\-te valu\-es from both compared data sets are drawn from the same distribution. We report the percentage of numeric attributes, where this hypothesis was rejected at 0.05 level (lower value indicates higher similarity). For discrete attributes we report average Hellinger distance over all discrete attribute in each data set. The response variables are excluded from the data sets for this comparison as the similarity of their distributions is enforced by the generator. 

\textit{2. Clustering:} the distance based structure of original and constructed data sets is compared with k-medoids clustering, using Adjusted Rand Index (ARI) \cite{Hubert85}.
The ARI has expected value of 0 for random distribution of clusters, and value 1 for perfectly matching clusterings; the value of ARI can also be negative.
The ARI is used to compare two different clusterings on the same set of instances, while we want to compare similarity of two different sets of instances. To overcome this obstacle, we cluster both data sets separately and extract medoids of the clusters for each clustering.
The medoid of a cluster is an existing instance in the cluster, whose average similarity to all instances in the cluster is maximal.
For each instance in the first data set, we find the nearest medoid in the second clustering and assign it to that cluster, thereby getting a joint clustering of both data sets based on the cluster structure of the second data set. We repeat the analogous procedure for the second data set and get a joint clustering based on the first data set. These two joint clusterings are defined on the same set of instances (union of both original and generated data), therefore we can use ARI to assess similarity of the clusterings and compare structure of both data sets. 
The response variables was excluded from the data sets. For some data sets the ARI exhibits high variance, so we report the average ARI over 100 repetitions of new data  generation.

\textit{3. Classification performance:} we compare a predictive similarity of the data sets using classification accuracy of random forests.
We selected random forests due to robust performance of this learning algorithm under various conditions \cite{Verikas11}.
	The implementation used comes from R package CORElearn \cite{CORElearn}. The default parameters are used: 100 random trees with the number of randomly selected attributes in nodes set to square root of the number of attributes.
	We report 5x2 cross-validated performance of models trained and tested separately on both data sets.

The basic idea of classification based comparison of data sets is to train models separately on the original and generated data set. Both models are tested on yet unseen original and generated data and the performances are compared. If the performance of a model trained on the original data is comparable for original and generated data this is an indicator that the generated data is within the original distribution (i.e., there are no significant outliers and all the aspects of the original data are captured). 
If performance of a model trained on the generated data is comparable for original and generated data, this shows that the generated data enables comparable learning and has a good coverage of the original distribution, therefore the generator is able to produce good substitute for original data concerning machine learning and data mining.
Additionally, if the model trained on the original data achieves better performance on the generated data than on original data, this indicates that the generator is oversimplified and does not cover all peculiarities of the original data. 

We start with two data sets d1 and d2 (e.g. original and generated one) and split them randomly but stratified into two halves (d1 produces d1a and d1b, d2 is split into d2a and d2b). Each of the four splits is used to train a classifier, and we name the resulting models m1a, m1b, m2a, and m2b, respectively. We evaluate the performance of these models on data unseen during training, so m1a is tested on d1b and d2, m1b is tested on d1a and d2, m2a uses d2b and d1, and m2b uses d2a and d1 as the testing sets. Each test produces a performance score (e.g., classification accuracy, AUC...), which we average as in a 2-fold cross-validation and get the following estimates:
\begin{enumerate}
\item performance of m1 on d1 (models built on original data and tested on original data) is an average performance of m1a on d1b and m1b on d1a,
\item performance of m1 on d2 (classifiers built on original data and tested on generated data) is an average performance of m1a on d2 and m1b on d2,
\item performance of m2 on d2 (models built on generated data and tested on generated data) is an average performance of m2a on d2b and m2b on d2a,
\item performance of m2 on d1 (models built on generated data and tested on original data) is an average performance of m2a on d1 and m2b on d1.
\end{enumerate}
These estimates already convey an important information as discussed above, but we can subtract performances 
of m1 on d1 and m2 on d1 (models built on original and generated data, both tested on original  data) to get $\Delta$d1. This difference, in our opinion, is the most important indicator of how suitable is the generated data for development of classification  methods. If, in ideal case, this difference is  close to zero, the generated data would be a good substitute for the lack of original data, and we can expect that performance of developed methods will be comparable when used on original data.
%

 \phantom{m} 

The results of all three types of comparisons are presented in  Table \ref{tab:w1b05} for binary encoding of nominal attributes, $minW=1$, and \textit{deafultSpread=}0.05.
These parameters form the default set of parameters and were selected with preliminary tests on artificial data. We tested also other sets of parameters,
 but the generated data sets were  less similar to the original data sets. The detailed results and the evolution of our approach is available in our technical report \cite{Robnik14rbfGenTR}.
\begin{table*}[t]
\caption{The comparison of original and generated data sets using the default set of parameters. 
}
\centerline{
\scriptsize
\begin{tabular}{lrrrrrrrrrrrrrr}
dataset  & $G$ & $t$ & $=$ & $m|\Delta mean|$ &  $m|\Delta std|$ & $m|\Delta \gamma_1|$ & $m|\Delta \gamma_2|$ & $KSp$ & $\overline{H}$
             & ARI &  m1d1 & m1d2 & m2d1 & m2d2 \\ \hline
             annealing & 143& 12.0 &   0 &  0.097 &  0.012  & 0.51 & 0.91  & 100 & 0.029& 0.721 & 98 &  92 &  97 &  93 \\
            arrhythmia & 421& 40.5 &   0 &  0.039 &  0.059  & 0.70 & 3.30  &  97 & 0.035& 0.218 & 71 &  69 &  56 &  96 \\
             audiology & 161&  8.3 &  36 &      - &      -  &    - &    -  &   - & 0.052& 0.174 & 71 &  83 &  71 &  83 \\
            automobile & 125&  4.7 &   0 &  0.062 &  0.062  & 0.36 & 1.24  &  53 & 0.061& 0.611 & 71 &  67 &  65 &  67 \\
         balance-scale & 212&  4.5 &   0 &  0.006 &  0.040  & 0.02 & 0.07  & 100 &     -& 0.288 & 84 &  85 &  83 &  88 \\
         breast-cancer & 208&  5.4 &  66 &      - &      -  &    - &    -  &   - & 0.059& 0.176 & 70 &  86 &  77 &  94 \\
    breast-cancer-wdbc & 151&  6.0 &   0 &  0.022 &  0.054  & 0.22 & 1.61  &  97 &     -& 0.951 & 95 &  97 &  95 &  98 \\
reast-cancer-wisconsin & 122&  3.9 &   0 &  0.008 &  0.031  & 0.07 & 0.19  & 100 &     -& 0.968 & 96 &  98 &  97 &  98 \\
      bridges.version1 &  68&  1.4 &   0 &  0.055 &  0.027  & 0.35 & 0.31  &  25 & 0.089& 0.726 & 63 &  74 &  70 &  83 \\
      bridges.version2 &  77&  1.9 &   0 &  0.014 &  0.002  & 0.35 & 0.02  &   0 & 0.073& 0.126 & 61 &  82 &  68 &  96 \\
                  bupa & 266&  4.4 &   0 &  0.054 &  0.061  & 0.76 & 4.95  & 100 &     -& 0.226 & 69 &  65 &  63 &  73 \\
      credit-screening & 342& 13.4 &   0 &  0.096 &  0.060  & 1.11 &12.17  &  83 & 0.032& 0.115 & 87 &  92 &  87 &  94 \\
        cylinder-bands & 355& 27.1 &   0 &  0.026 &  0.071  & 0.42 & 1.65  & 100 & 0.108& 0.602 & 78 &  73 &  73 &  86 \\
           dermatology & 257& 15.6 &   0 &  0.026 &  0.051  & 0.03 & 0.19  &   0 & 0.046& 0.914 & 97 &  97 &  96 &  98 \\
                 ecoli & 131&  2.8 &   0 &  0.024 &  0.034  & 0.16 & 0.52  &  57 &     -& 0.810 & 84 &  85 &  85 &  86 \\
                 flags & 167&  8.0 &   0 &  0.173 &  0.077  & 6.08 &63.83  & 100 & 0.109& 0.218 & 61 &  74 &  57 &  97 \\
                 glass & 123&  2.3 &   0 &  0.036 &  0.043  & 0.55 & 2.63  &  67 &     -& 0.383 & 72 &  63 &  69 &  77 \\
              haberman & 149&  3.1 &   0 &  0.059 &  0.046  & 0.78 & 4.70  &  50 & 0.055& 0.817 & 71 &  75 &  76 &  80 \\
eart-disease-cleveland & 217&  4.9 &   0 &  0.036 &  0.052  & 0.12 & 0.34  &  83 & 0.043& 0.537 & 57 &  72 &  61 &  96 \\
eart-disease-hungarian & 124&  2.8 &   0 &  0.014 &  0.052  & 0.24 & 1.16  &  67 & 0.147& 0.751 & 82 &  91 &  83 &  93 \\
             hepatitis &  83&  2.0 &   0 &  0.027 &  0.043  & 0.43 & 0.65  &   0 & 0.058& 0.151 & 84 &  91 &  84 &  97 \\
           horse-colic & 314& 12.5 &  84 &  0.044 &  0.037  & 0.39 & 0.67  &  57 & 0.123& 0.323 & 84 &  91 &  85 &  93 \\
        house-votes-84 & 180&  6.1 &  53 &      - &      -  &    - &    -  &   - & 0.030& 0.840 & 96 &  99 &  94 &  99 \\
            ionosphere & 160&  5.2 &   0 &  0.011 &  0.009  & 0.09 & 0.49  &  79 &     -& 0.540 & 93 &  97 &  85 &  98 \\
                  iris &  23&  1.5 &   0 &  0.005 &  0.018  & 0.10 & 0.11  &  25 &     -& 0.897 & 95 &  95 &  96 &  98 \\
    labor-negotiations &  37&  0.8 &   0 &  0.086 &  0.040  & 0.44 & 1.60  &  38 & 0.208& 0.048 & 86 &  86 &  88 &  87 \\
          lymphography & 117&  2.7 &   0 &  0.089 &  0.030  & 0.30 & 0.45  & 100 & 0.086& 0.284 & 80 &  90 &  79 &  93 \\
               monks-1 & 184&  4.9 &  96 &      - &      -  &    - &    -  &   - & 0.015& 0.060 & 97 &  95 &  97 &  93 \\
               monks-2 & 345&  8.1 &  94 &      - &      -  &    - &    -  &   - & 0.008& 0.114 & 80 &  88 &  79 &  82 \\
               monks-3 & 207&  5.5 &  96 &      - &      -  &    - &    -  &   - & 0.011& 0.091 & 98 &  96 &  98 &  95 \\
 pima-indians-diabetes & 506& 12.3 &   0 &  0.054 &  0.041  & 0.69 & 2.32  & 100 &     -& 0.295 & 76 &  81 &  76 &  84 \\
        post-operative &  68&  1.0 &   4 &  0.044 &  0.046  & 0.86 & 1.27  & 100 & 0.102& 0.422 & 66 &  76 &  70 &  88 \\
         primary-tumor & 288&  8.2 &  79 &      - &      -  &    - &    -  &   - & 0.058& 0.109 & 45 &  71 &  44 &  88 \\
             promoters & 101&  7.5 &   3 &      - &      -  &    - &    -  &   - & 0.335& 0.233 & 88 &  98 &  50 & 100 \\
             sonar.all & 128&  5.2 &   0 &  0.015 &  0.033  & 0.23 & 0.65  &  18 &     -& 0.236 & 78 &  91 &  82 &  95 \\
         soybean-large & 300& 41.6 &  28 &      - &      -  &    - &    -  &   - & 0.058& 0.385 & 92 &  94 &  83 &  96 \\
           spect-SPECT & 161&  3.8 &  57 &      - &      -  &    - &    -  &   - & 0.020& 0.836 & 82 &  89 &  85 &  89 \\
          spect-SPECTF & 215&  5.7 &   0 &  0.030 &  0.079  & 0.87 & 3.58  & 100 &     -& 0.203 & 81 & 100 &  79 & 100 \\
          spectrometer & 473& 33.1 &   0 &  0.019 &  0.082  & 0.90 & 4.01  &  97 &     -& 0.969 & 49 &  57 &  38 &  88 \\
                sponge &  44&  3.7 &  10 &      - &      -  &    - &    -  &   - & 0.081& 0.966 & 92 &  97 &  93 &  97 \\
    statlog-australian & 333&  9.6 &   0 &  0.101 &  0.071  & 1.25 &12.77  & 100 & 0.030& 0.939 & 87 &  91 &  87 &  93 \\
        statlog-german & 845& 39.1 &   0 &  0.077 &  0.036  & 0.30 & 0.86  & 100 & 0.060& 0.253 & 75 &  83 &  76 &  96 \\
statlog-german-numeric & 723& 24.5 &   0 &  0.043 &  0.038  & 0.22 & 0.34  & 100 &     -& 0.159 & 75 &  80 &  76 &  83 \\
         statlog-heart & 139&  2.7 &   0 &  0.045 &  0.050  & 0.15 & 0.43  &  75 & 0.056& 0.667 & 81 &  92 &  83 &  94 \\
       statlog-vehicle & 571& 17.5 &   0 &  0.031 &  0.027  & 0.13 & 0.16  & 100 &     -& 0.954 & 74 &  72 &  70 &  80 \\
                   tae &  90&  1.0 &   0 &  0.042 &  0.018  & 0.20 & 0.08  &  33 & 0.037& 0.350 & 50 &  54 &  54 &  55 \\
   thyroid-disease-new &  38&  1.3 &   0 &  0.031 &  0.022  & 0.48 & 4.15  &  60 &     -& 0.497 & 96 &  85 &  91 &  90 \\
           tic-tac-toe & 859& 28.6 &  78 &      - &      -  &    - &    -  &   - & 0.090& 0.164 & 95 &  98 &  74 &  99 \\
         vowel-context & 291& 12.4 &   0 &  0.004 &  0.016  & 0.06 & 0.08  &   0 &     -& 0.441 & 88 &  77 &  85 &  76 \\
                  wine &  52&  1.8 &   0 &  0.023 &  0.037  & 0.10 & 0.28  &  31 &     -& 0.487 & 97 &  96 &  97 &  95 \\
                   zoo &  23&  1.4 &   0 &  0.010 &  0.003  & 0.37 & 0.02  & 100 & 0.025& 0.575 & 91 &  94 &  94 &  94 \\
\hline
\end{tabular}
}
\label{tab:w1b05}
\end{table*}

The column labeled  with $G$ shows the number of Gaussian kernels in each of the constructed generators.
 Relatively large numbers of units are needed to adequately represent the
training data. Nevertheless, the time needed to construct the generators (with function \textit{rbfDataGen}) is low as seen from the column labeled with $t$, which gives times in seconds. For measurements we used a single core of Intel i7 CPU running at 2.67GHz, 12GB  RAM, Windows 7, and R 3.1.0. The time to generate the data (with function \textit{newdata}) was below 1 sec for 1000 instances in all cases, so we do not report it. 

The column labeled with $=$ gives the percentage of generated instances exactly equal to the original instances. This mostly happens in data sets with only discrete attributes where the whole problem space is small and identical instances are to be expected.
Exception from this are datasets horse-colic, primary-tumor, and breast-cancer, where majority of Gaussian units  in generators contain only one 
activation instance. The reason for this is large number of attributes and consequently a poor generalization in \textit{rbfDDA} algorithm.

Columns  labeled $m|\Delta mean|$  and  $m|\Delta std|$ report median of absolute values of differences in means and standard deviations for attributes normalized to $[0, 1]$. In 37 (39) out of 39 cases the absolute differences in means (standard deviations) are below $0.10$. In 27 (25) out of 39 cases the differences are below $0.05$.  
These small differences in means and standard deviations indicate that first and second central moment of the generated distributions for individual attributes are close to the originals. 

Columns  labeled $m|\Delta \gamma_1|$ and $m|\Delta \gamma_2|$ report median of absolute values of differences in skewness and kurtosis, respectively. These two indicators require symmetric distribution for interpretation,  are sensitive to outliers, and their values are unrestricted (the span is from $-\infty$ to $\infty$) \cite{Brys06}. These properties make them difficult to summarize and compare across several attributes. We observe that for some data sets the indicators
report low differences in skewness (and for a lesser extent also for kurtosis), while for others the differences are substantial. We tried alternative measures for skewness and kurtosis proposed in \cite{Brys06}, but the results were similar. Closer inspection of individual data sets revealed that it is not possible to make general conclusions as the range of differences even for attributes within the same data set are large. The user is therefore advised to check these values with some knowledge of the domain. 

The distributions
of individual attributes are compared with KS test for numeric attributes and with Hellinger distance for discrete attributes.
Column labeled $KSp$ gives a percentage of p-values below 0.05 in KS tests comparing matching numeric attributes using the null hypothesis
that original and generated data are drawn from the same distribution. For most of the data sets and most of the attributes the KS-test detects the differences in distributions. Column labeled $\overline{H}$ presents average Hellinger distance for matching discrete attributes. While for many data sets the distances are low, there are also some data sets where the distances are relatively high, indicating that distribution differences can be considerable for discrete attributes.
 
The suitability of the generator as a development, simulation, or benchmarking tools in data mining is evidenced by comparing clustering and classification performance. The column labeled ARI presents adjusted Rand index. We can observe that clustering similarities are considerable for many data sets (high ARI) but there are also some data sets, where it is low. 

The columns m1d1, m1d2, m2d1, and m2d2 report 5x2 cross-validated classification accuracy of random forest models trained on either original (m1) or generated (m2) data and tested on both original (d1) and generated (d2) data. A general trend observed is that
on majority of data sets models trained on original data (m1) perform slightly better on generated data than on the original data
(accuracy m1d2 is higher than m1d1 for 22 of 51 data sets, for 20 data sets m1d2 is lower, and on 9 data sets there is a draw). This indicates that at least some of the complexity of the original data is lost in the generated data. This is confirmed also by models built on the generated data (m2) which mostly perform better on the generated than on the original data
(accuracy m2d2 is higher than m2d1 in 42 out of 51 data sets, in 7 cases m2d2 is lower, and in 2 cases there is a draw). Nevertheless, in many cases the generated data can be a satisfactory substitute in data mining development, namely models build on the generated data outperform models built on the original data when both are tested on original data 
($\Delta d_1<0$). In  20 cases out of 51 cases m2d1 is higher than m1d1, in 22 cases m1d1 is lower and there are 9 draws.

An overall conclusion is therefore that for a considerable number of data sets the proposed generator can generate semi-artificial data set which are reasonable substitutes in development of data mining algorithms. 

\subsection{When RBF-based generator works?}
\begin{table*}[!!ht]
\caption{The correlation coefficients between different factors than might influence the quality of RBF-based data generators and $\Delta d_1$.
}
\centerline{
\normalsize
\begin{tabular}{lrrrrrrrrrr}
 & $RBF$ & $RF$  & $\Delta acc$  &  $n$ & $a$ & $n/a$
             & $G$ &  $n/G$ & $a/G$  \\ \hline
$\Delta d_1$ &  -0.12  & 0.12 & 0.45 & 0.16 & 0.38 & -0.01 & 0.27 & -0.12 &  0.21 \\
\end{tabular}
}
\label{tab:correlation}
\end{table*}

We want to determine the conditions when RBF based data generation works well. We approach this problem by defining a success criterion for data generators and identification of different factors that could affect generators' performance.
As a measure of success we use the difference in classification accuracy of models trained on original data (m1) and generated data (m2) and tested on original data (d1). This difference is labeled $\Delta d_1$. Negative values of $\Delta d_1$ show that generated data are good substitute for original data. 

The first  group of success prediction factors is based on the success of RBF classification algorithm. As classification performance expressed in absolute numbers is not useful, we compared the classification performance of \textit{rbfDDA} with  performance of random forests.  We selected random forests as it is one of the most successful classifiers, known for its  robust performance (see for example \cite{Verikas11}). Using 5x2 cross-validation we computed  
the classification accuracy and AUC of \textit{rbfDDA} algorithm from RSNNS package \cite{Bergmeir12} and random forests from CORElearn package \cite{CORElearn}, using the default parameters for both classifiers. 
Unsurprisingly, random forest perform better than RBF networks. The difference in classification accuracy is significant at 0.05 level for 34 of 51 data sets (for 3 data sets RBF is significantly better, other differences are insignificant).

The prediction  factors extracted from this comparison are classification accuracies of RBF and RF, and the difference in classification accuracies between these two classifiers $\Delta acc$.
Other predictive factors used are the number of instances $n$,  number of attributes $a$, number of instances per attribute $n/a$,
 number of Gaussian units in the  generator $G$,  average number of instances per Gaussian kernel $n/G$, and number of attributes per Gaussian kernel $a/G$.  
In Table \ref{tab:correlation} we show their correlation with $\Delta d_1$.

The Pearson's correlation coefficients indicate largest correlation of performance with the difference in classification accuracy between RBF and RF, number of attributes, and number of Gaussian kernels. All these factors indicate difficulty of the problem for RBF classifier, which, in our opinion, hints that the usability of proposed generator depends on the ability of its learning method to capture the structure of the problem.

We tried to predict the success of the data generator using stepwise linear model with independent variables as above, but it turned out that  difference in classification accuracy between RBF and RF ($\Delta acc$) is the only variable needed. 

\subsection{Can semi-artificial data improve classification accuracy?}
\label{sec:accuracy}
In some experimental studies generated data are merged with the original data in order to improve the classification accuracy. For example, \cite{Li06} adds the data generated with low-dimensional Gaussian kernels to the original data set and reports improved classification accuracy of neural networks for a specific manufacturing data set on condition that just the right amount of new instances are added. In our opinion this could result from the fact that the hypothesis language of Gaussian kernels is different from the hypothesis language of neural networks therefore Gaussian kernels might capture different dependencies than neural networks. As the data set used in the reported experiment was very small, adding just the right amount of data where these dependencies were clearly expressed helped neural networks to learn better and resulted in better classification accuracy. 

Niyogi et al \cite{Niyogi98} show on a specific pattern recognition task  that adding virtual examples is equivalent to incorporating prior knowledge into the learning task. 
This result hints that it is unlikely that any general approach would be able to generate new instances which would improve the classification performance. Nevertheless, improved classification performance can more likely be expected, when the learning algorithms is different from data generator in a sense that their hypothesis description languages are different.

To test if new, generated instances, added to the training set significantly improve the classification accuracy we used the following scenario. We split a given data set into two halves and use each subset to construct a generator. We use the generator to produce \textit{the same number of new instances} as in the original subset. The generated data was merged with the original data subset and used as a training set for two learning algorithms (decision trees and random forest). The performance of the resulting classifiers was evaluated on yet unseen second halves of the data sets. The same learning algorithms were used also on the original data subset.
The above procedure was repeated by switching the training and testing subsets. The whole process was repeated 5 times and imitating Dietterich's 5x2 cross-validation test \cite{Dietterich98} we compared the statistical differences between the two classifiers, one trained on just the original data set and the other using both original and generated data. 

For random forests adding generated data significantly improves the classification accuracy on  5 data sets out of 51 (audiology,    iris, monks-2, spectrometer, and zoo),  The differences are significant at 0.05 level. For decision trees the differences are significant  for four data sets (soybean-large, vowel-context, wine, and zoo).
These results motivated us for further investigation of this research path and adding different amounts of generated instances.

We performed an analysis on the same 51 data sets used above. For five times we repeated the following procedure. We split a data set randomly into halves, construct a generator on one half of instances, generate 2000 new instances, add a certain amount of generated instances to the same training half of instances, train a classifier on the merged data, and test the classifier on the second half of the original data set. We repeat the same procedure on the second half of the data set and average the results. The number of added instances varied from 0 to 2000 (more specifically from 0 to 200 in steps of 10, from 200 to 500 in steps of 50, and from 500 to 2000 in steps of 100 instances). As classifiers we used random forests and decision trees. The results of 5 x 2 repetitions were averaged.

\begin{figure*}[!!tp]
\centerline{\includegraphics[width= 0.49 \hsize, height=0.21 \vsize ]{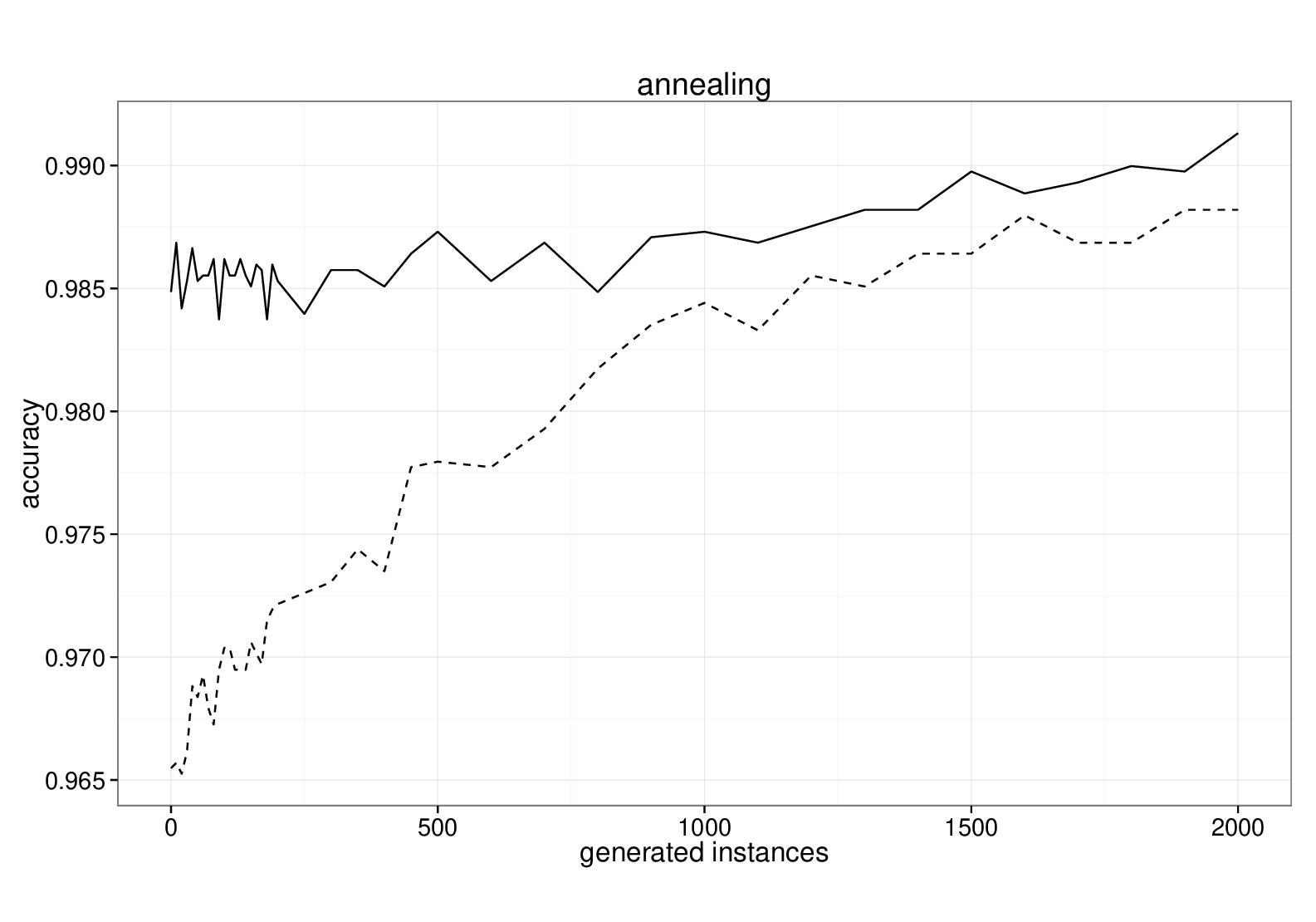} \
                 \includegraphics[width= 0.49 \hsize, height=0.21 \vsize ]{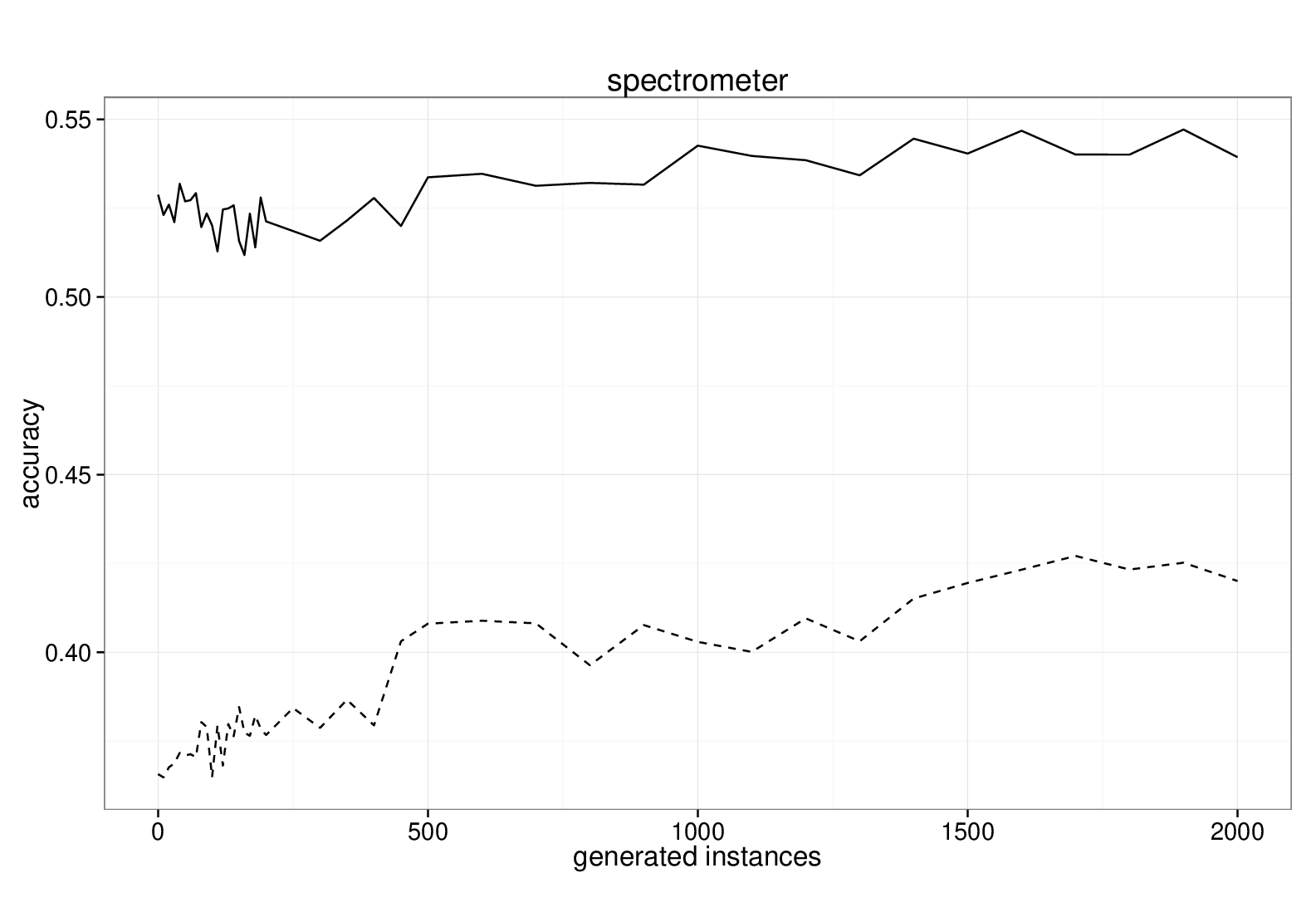} } 
\centerline{\hskip4.5cm a) \hfill  b) \hskip4cm }
\centerline{\includegraphics[width= 0.49 \hsize, height=0.21 \vsize ]{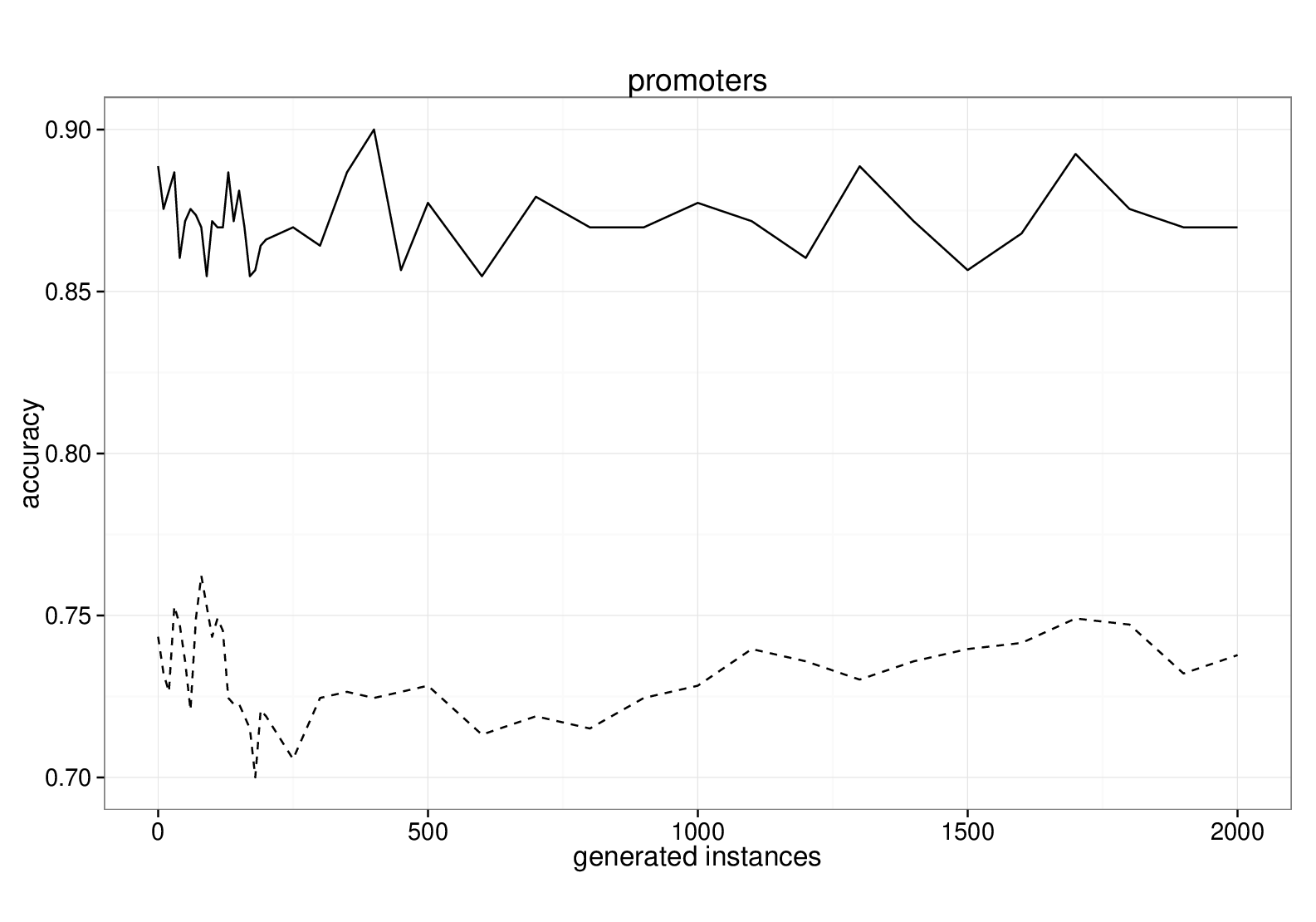}
                 \includegraphics[width= 0.49 \hsize, height=0.21 \vsize ]{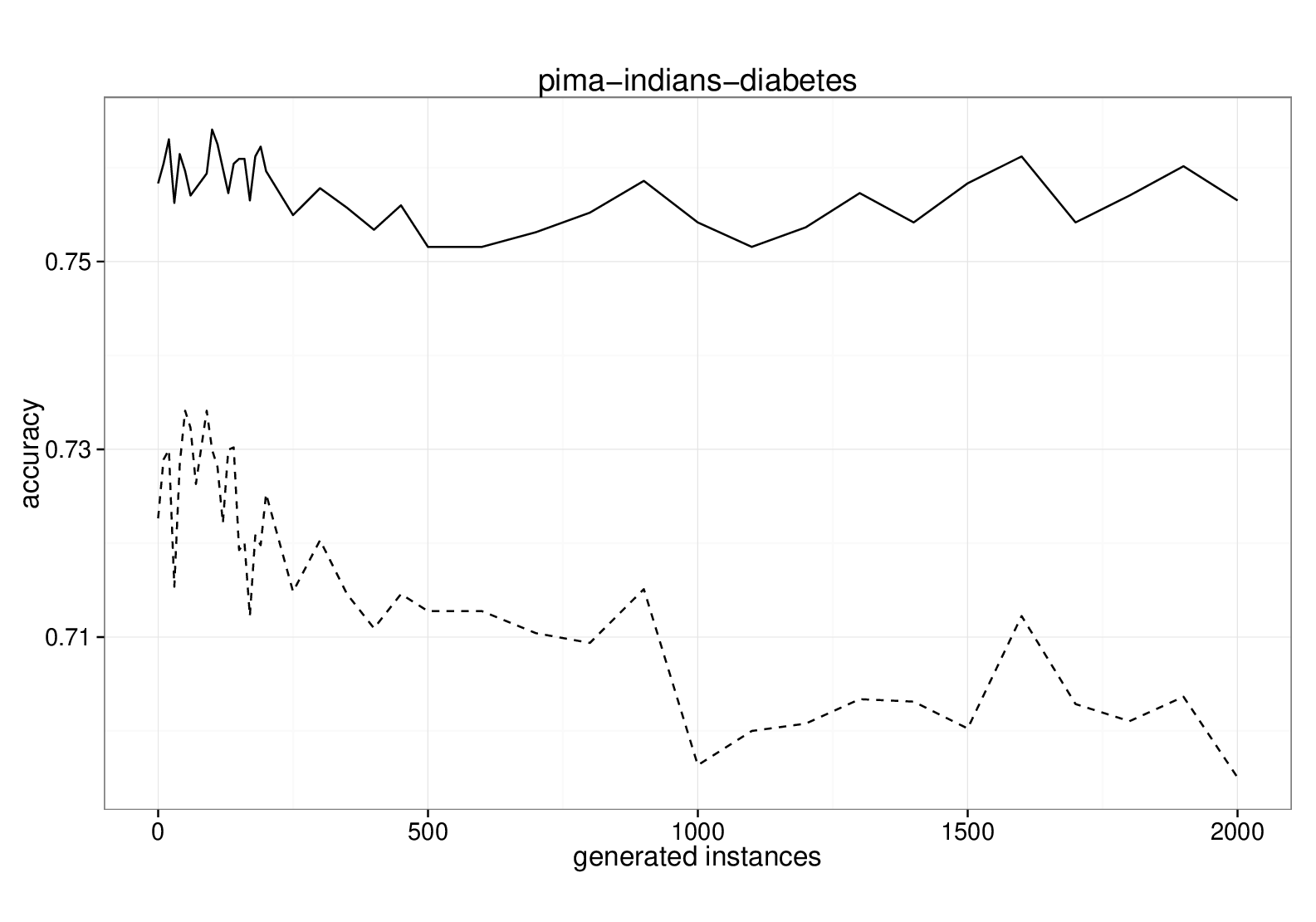} }
\centerline{\hskip4.5cm c) \hfill  d) \hskip4cm }
\centerline{\includegraphics[width= 0.49 \hsize, height=0.21 \vsize ]{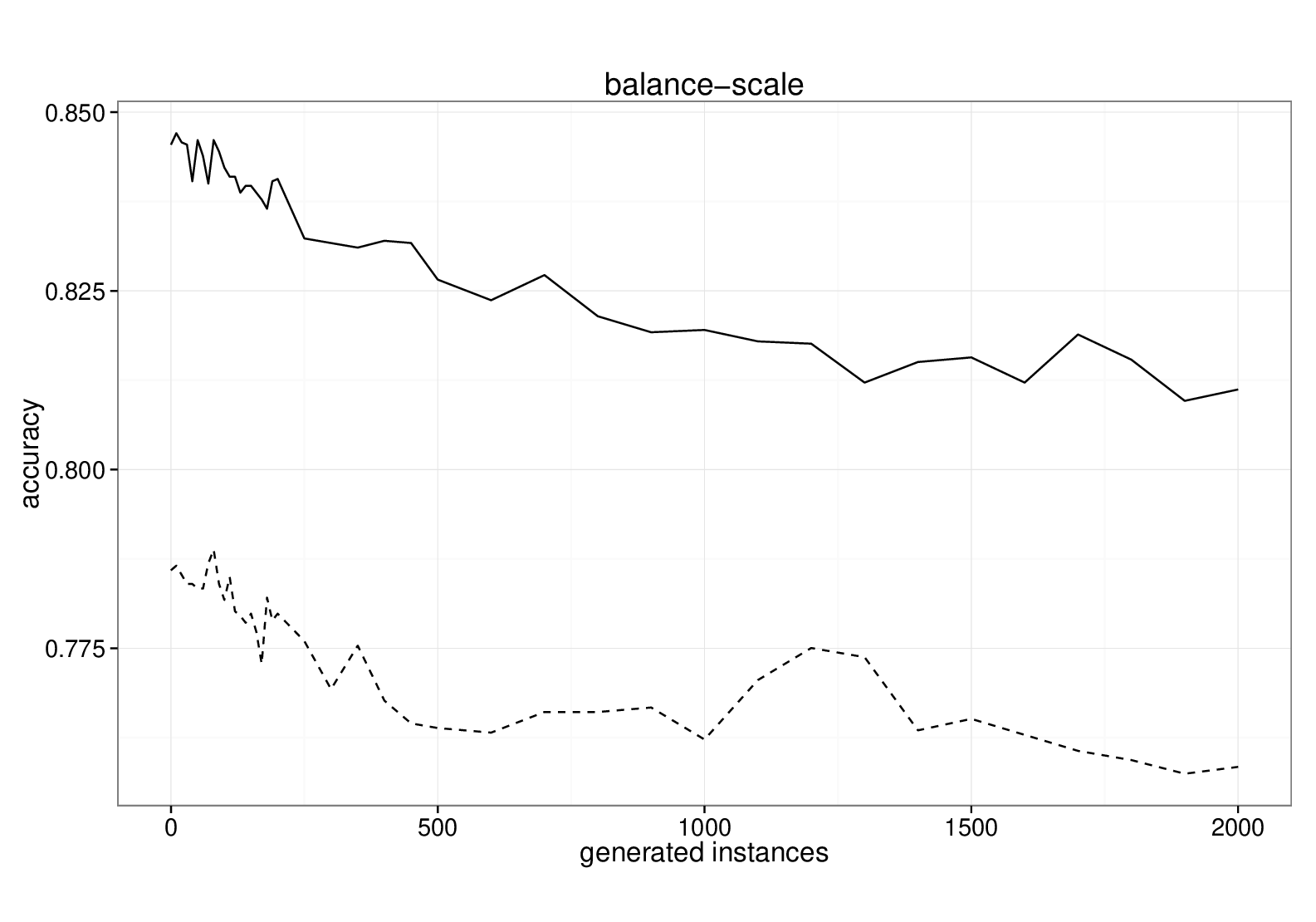}
                  \includegraphics[width= 0.49 \hsize, height=0.21 \vsize ]{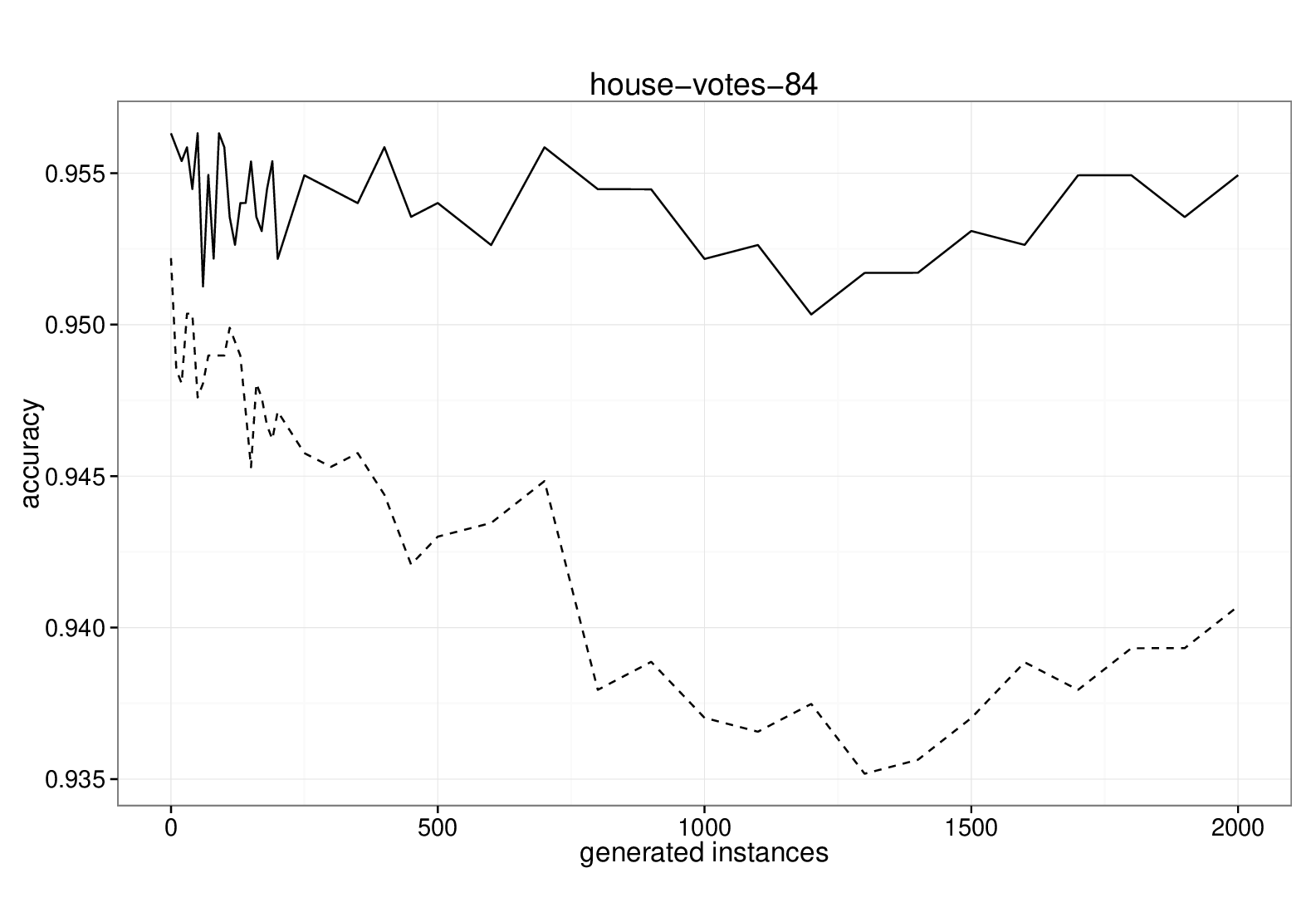} }
\centerline{\hskip4.5cm e) \hfill  f) \hskip4cm }
\centerline{\includegraphics[width= 0.49 \hsize, height=0.21 \vsize ]{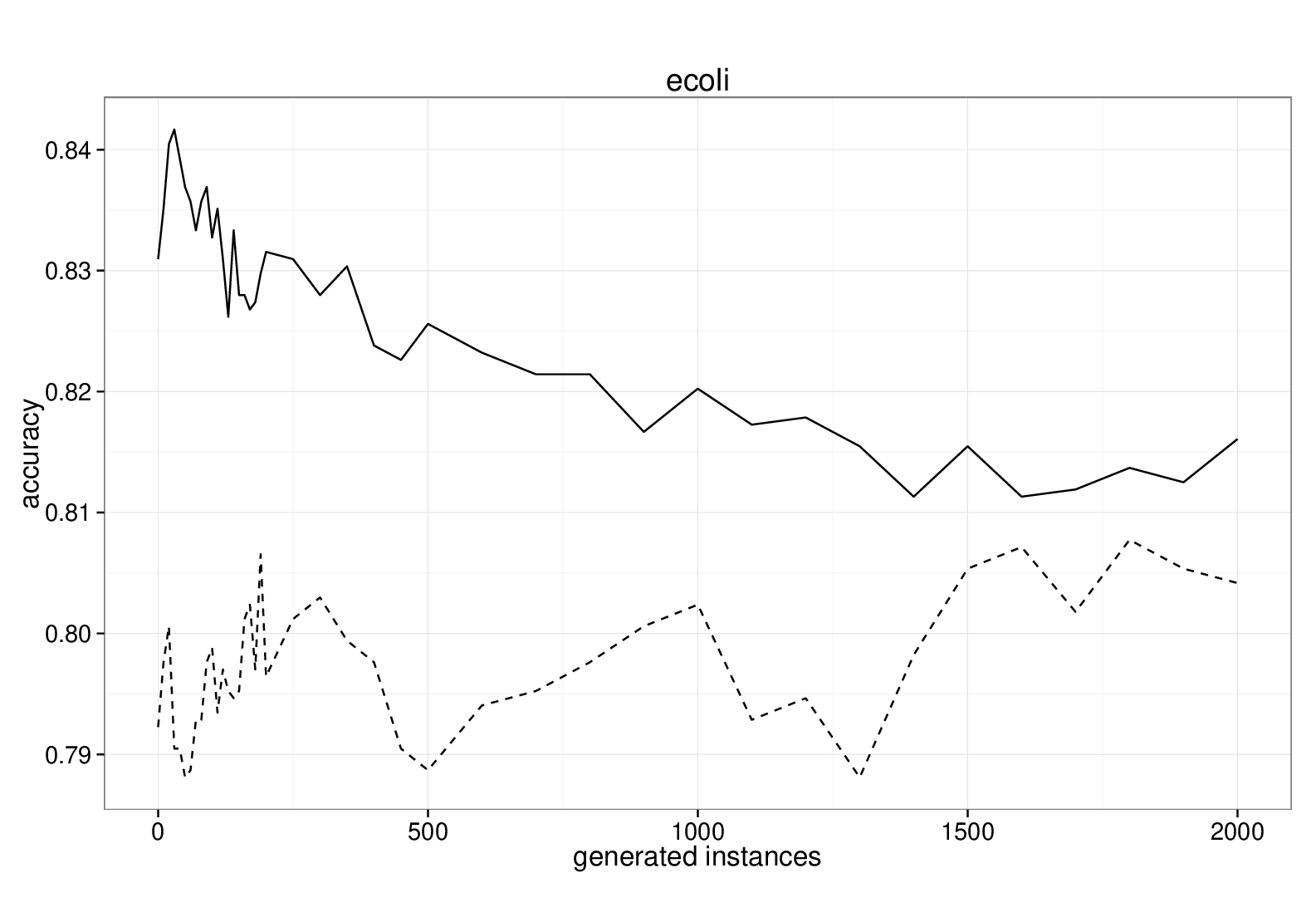}
                 \includegraphics[width= 0.49 \hsize, height=0.21 \vsize ]{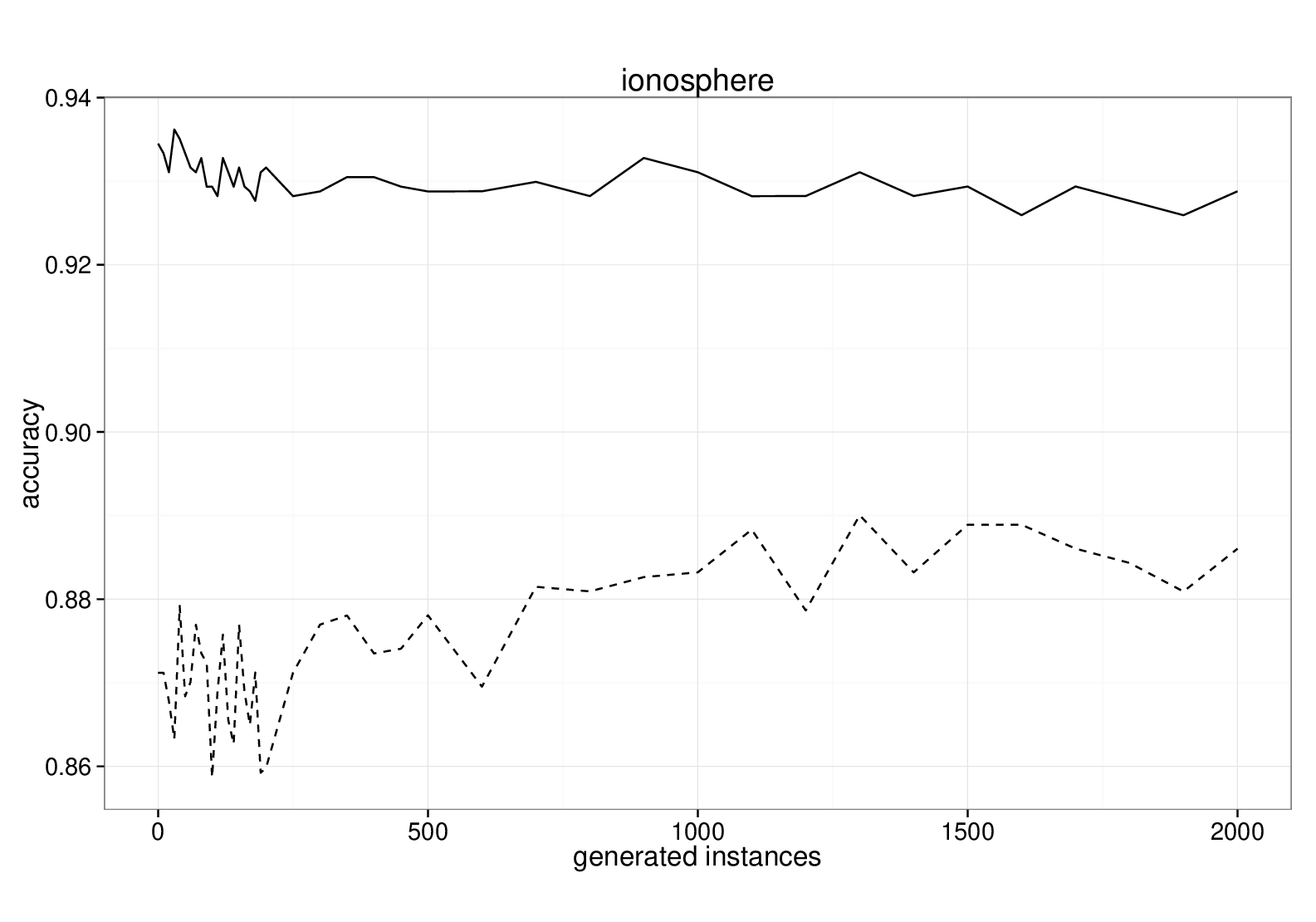} }
\centerline{\hskip4.5cm g) \hfill  h) \hskip4cm }
\caption{Effect of added generated data on classification accuracy for different data sets. Solid and dashed lines represents classification accuracy of random forests and decision trees, respectively. The accuracy scale is different for each graph. Plots a) -- h) show results for data sets
annealing, spectrometer, promoters, pima-indians-diabetes, balance-scale, house-votes-84, ecoli, and ionosphere, respectively.} 
 \label{fig:genSizes}
\end{figure*}
The results show that the number of added instances is an important, but data set specific, factor. In Fig. \ref{fig:genSizes} we present some typical plots depicting the dependence of classification accuracy on the number of added generated instances. Each plot contains two lines: solid lines show behavior of random forest classifiers, and dashed lines present decision tree classifiers.
Our analysis shows three types of typical behavior:
\begin{itemize}
\item accuracy improves with more generated data and gradually levels out (random forests: 10 data sets, decision tree: 22 data sets with this type of behavior), examples of such behavior are  data sets annealing and spectrometer in Figs. \ref{fig:genSizes}a and \ref{fig:genSizes}b;
\item some (mostly small) amount of added generated data improves classification accuracy, but with more data the accuracy decreases or stays the same (random forests: 17 data sets, decision tree: 17 data sets with this type of behavior), examples of such behavior are  data sets promoters and pima-indian-diabetes in Figs. \ref{fig:genSizes}c and \ref{fig:genSizes}d;
\item adding generated data decreases the classification accuracy (random forest: 24 data sets, decision tree: 12 data sets with this type of behavior),
examples of such behavior are  data sets balance-scale and house-votes-84 in Figs. \ref{fig:genSizes}e and \ref{fig:genSizes}f;
\end{itemize}
The behaviors are sometimes also classifier dependent as Figs. \ref{fig:genSizes}g and \ref{fig:genSizes}h illustrate, where 
on data sets ecoli and ionosphere small amount of generated data improve classification accuracy of random forests, but 
for decision trees accuracy further improves with adding more generated data.
The fact that classification accuracy can be improved for more data sets using decision tree classifier than random forests, seems to support our speculation that different hypothesis description language of RBF networks is a possible cause for improvements in  classification accuracy. The description language of decision trees is much more restricted than the description language of random forests, so adding more instances has greater impact on decision trees.

An overall conclusion is that getting statistically significant improvements is possible by adding generated data to the training set, but for most data sets classification accuracy improves only if the right amount of generated data is added. The number of instances required can be extracted from graphs similar to the ones in Fig. \ref{fig:genSizes}.

\subsection{High dimensional data sets}
\begin{figure*}[!!t]
\centerline{
\includegraphics[width= 0.49 \hsize, height=0.21 \vsize]{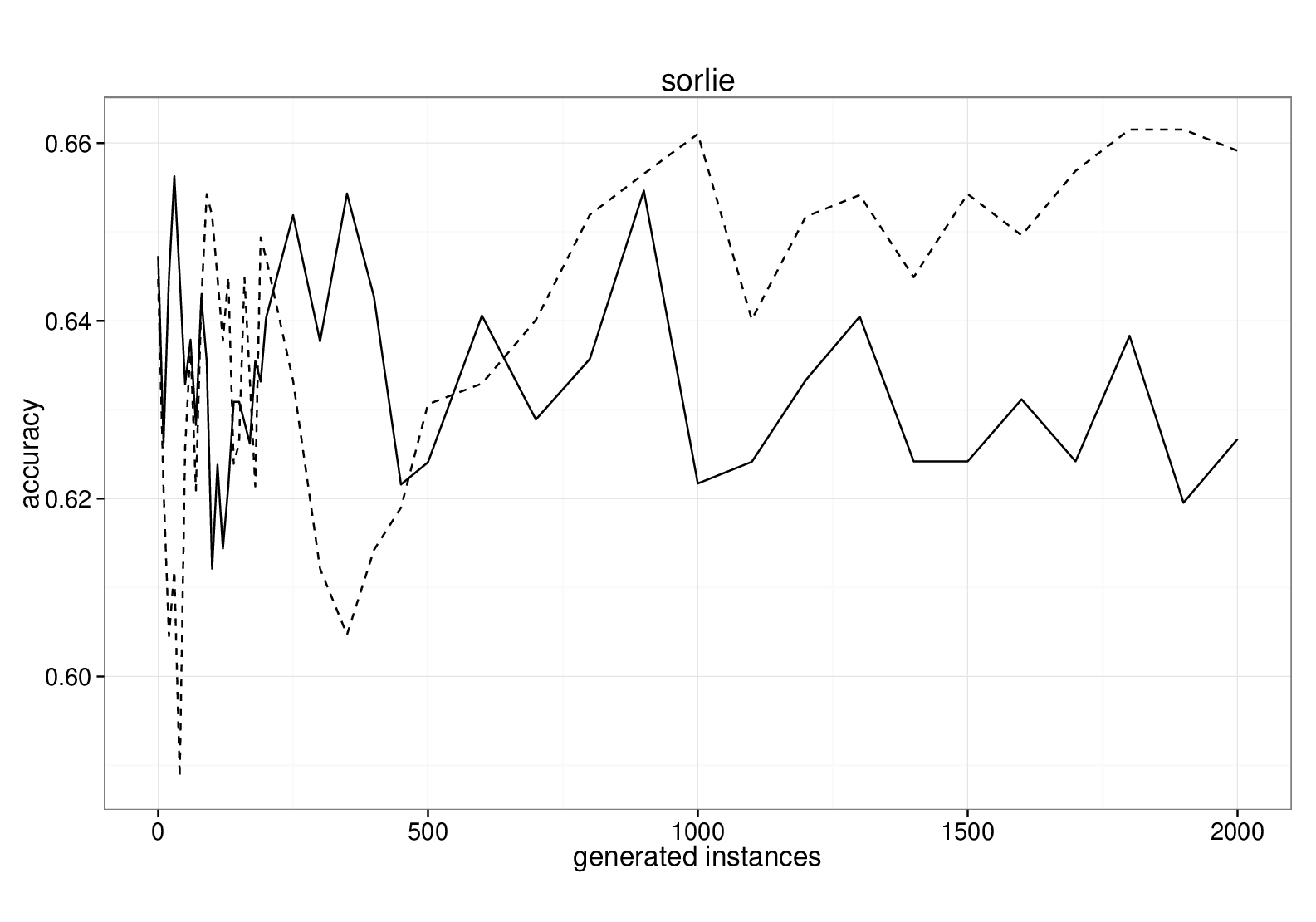}
\includegraphics[ width= 0.49 \hsize, height=0.21 \vsize]{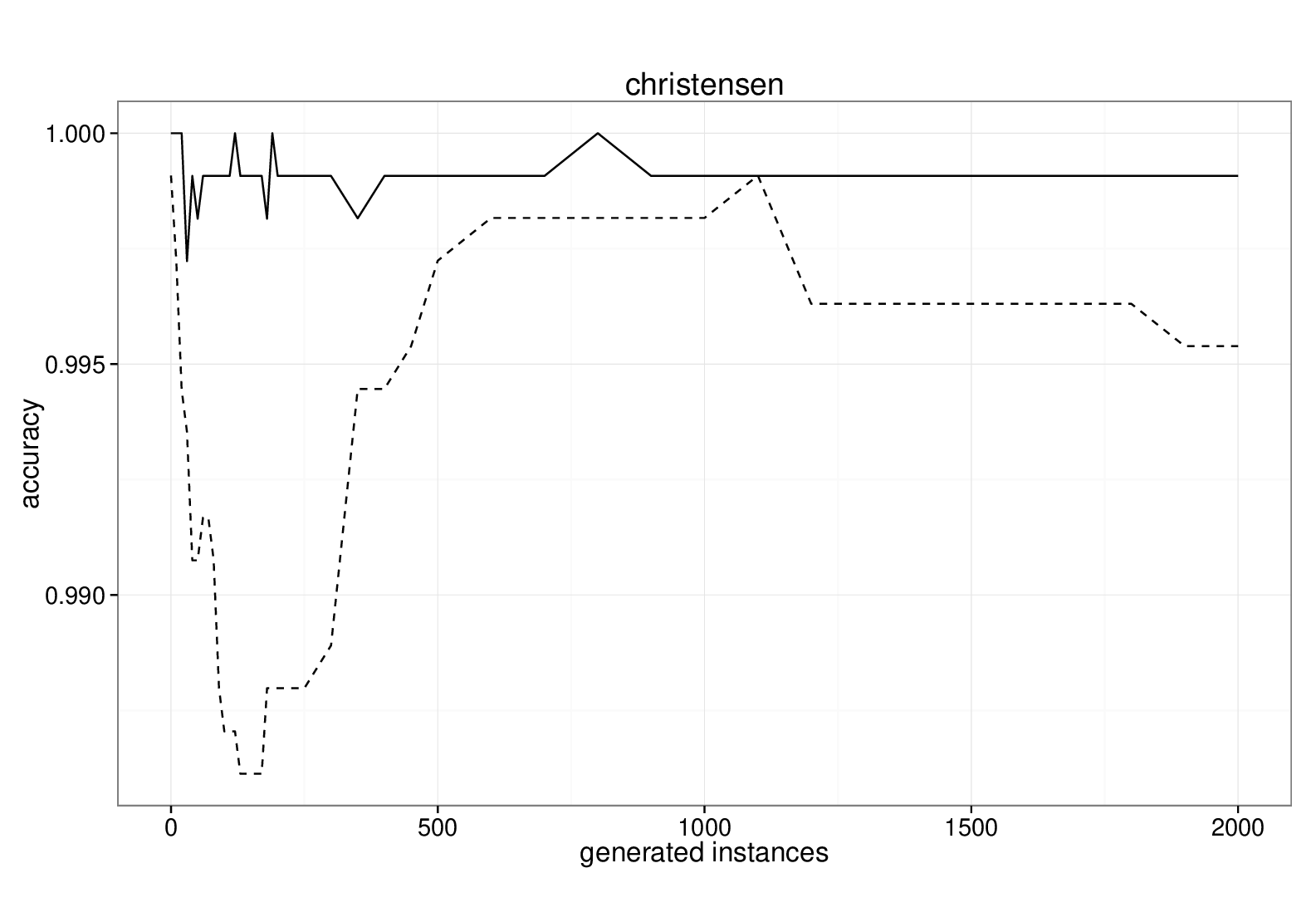} 
}
\caption{Effect of generated data on classification accuracy for two gene expression data sets - sorlie and christensen. Solid and dotted lines represents classification accuracy of random forests and decision trees, respectively. The accuracy scale is different for each graph.} 
 \label{fig:microarraySize}
\end{figure*}

Many applications in bioinformatics suffer from the shortage of data for modeling or classification. Semi-artificial data could be very useful in this area. We discuss behavior of the proposed data generator on these type of data using as an example two data sets from the R datamicroarray package \cite{Rdatamicroarray} (sorlie and christensen). The sorlie data set consists of 85 instances described with 456 numeric attributes and christensen consists of 217 instances described with 1413 numeric attributes. We tested these two data sets with the same setting as in Section \ref{sec:accuracy}. The graphs on Fig. \ref{fig:microarraySize} show the effect of the number of generated data added to the original training set on the classification accuracy. We see that there is no improvement for random forest, but the accuracy of decision tree classifier improves with more added instances on sorlie data set (left-hand side) and even exceeds the accuracy of random forests. 

While this result indicates that semi-artificial data can be beneficial also for high dimensional data sets, data generators based on Gaussian kernels may not be an adequate solution. The function \textit{newdata} (Fig. \ref{alg:rbfNewdata}) requires matrix decomposition and multiplication involving the covariance matrix (see Equations (\ref{eq:linN}) and (\ref{eq:newN})). For data set with $a$ attributes the matrices involved in computation contain $a \times a$ elements. It is not uncommon for microarray data to contain tens of thousands of attributes, so the proposed approach would require enormous amount of memory (for a data sets with 50,000 attributes the amount of memory required to store a single matrix exceeds 18GB). For sorlie and christensen data sets  the generator construction takes 11 and 74 seconds, respectively, but  the generation of 1,000 instances takes 327 and 4,412 seconds, respectively. 

To tackle really high dimensional data sets (with more than e.g., 1,000 attributes) the proposed approach is therefore inadequate. We plan to solve this problem by turning learning algorithms not based on Gaussian kernels into data generators.

\subsection{Development of big data tools}
During the development of big data cloud based framework ClowdFlows \cite{Kranjc14}, we tested several classification algorithms which are components of the framework and also the capabilities and scalability of the framework. While big data community has produced several public data sets, each requires (sometimes tedious) preprocessing and adaptations to the specifics of the problem. As the development already  required a significant effort of everyone involved, such an additional effort was undesired. The use of function \textit{rbfDataGen} contained in an open-source R package semiArtificial \cite{semiArtificial} turned out to require little additional work but provided required benchmarking data. We generated several data sets with different characteristics (varying the number of features, instances, and proportions of classes), which  were needed during the development and evaluation of the framework. The use of the proposed generator turned out to be a good choice.

\section{Conclusions and further work}
We present an original and practically useful generator of semi-artificial data.
The generator captures structure of the problem using RBF classifier and exploits properties of Gaussian kernels to generate new data similar to the original one. We solved several practical issues and produced an efficient generator, which,  together with statistical, clustering and classification performance indicators, is available in an open-source R package semiArtificial \cite{semiArtificial}.  The generator was successfully tested in variety of settings including development of big data toolkit. 

The reasons for relative success of the proposed generator are both theoretical and practical. From theoretical point of view we do not assume that the whole data set comes from the same distribution, but use RBF networks to learn local probability density approximations. Additionally, in our implementation we solve many practical details: we encode multivalued nominal attributes with several binary attributes and can therefore handle mixed types of attributes, we only retain Gaussian kernels with sufficient statistical support, empirically estimate variance in each dimension, impute missing values, enforce consistency of generated instances, and optionally generate new instances according to an arbitrary class distribution.

We expect that semi-artificial data prepared with our method is going to be useful in the development and adaptation of data analytics tools to specifics of data sets. Possible other uses are data randomization  
to ensure privacy, simulations requiring large amounts of data, testing of big data tools, benchmarking, and scenarios with huge amounts of data.

Several data set similarity evaluation methods described and implemented in this work can provide an estimate of generator's performance for a specific data set. Using a large collection of UCI data sets we showed that the generator is in many cases successful in generating artificial data similar to the original. Furthermore, for some data sets adding a certain proportion of generated data to the training set significantly improves the classification accuracy. The success of individual generators is related to the success of RBF classifiers on the same data set. If RBF classifier can successfully  capture the properties of the original data,  the generator based on it is also going to be successful in most cases.
Still, we were unable to create a successful prediction model for the quality of the generator. The user is advised to use the provided data similarity evaluation tools. The results shall provide a good indication on the usability of the generated data for the intended use. 

In future work we plan to extend the generator with new modules using other learning algorithms 
to capture data structure and generate new data. An interesting path would also be a rejection approach which uses 
probability density estimates captured from various learning algorithms.

\section*{Acknowledgments}
\addcontentsline{toc}{section}{Acknowledgment}
Many thanks to Petr Savicky for interesting discussions on the subject and for preparation of UCI data sets used in this paper. The comments of anonymous reviewers and editors raised new questions and helped to significantly improve the paper. 

\bibliography{IEEEabrv,learn,my,util}

\end{document}